%% file: iclr2027_conference.tex
\documentclass{article} 
\usepackage{iclr2027_conference,times}

\input{math_commands.tex}

\usepackage{hyperref}
\usepackage{url}

\usepackage{graphicx}
\usepackage{amsmath,amssymb,amsthm}
\newtheorem{theorem}{Theorem}
\usepackage{booktabs}
\usepackage[table]{xcolor}
\usepackage{array}
\usepackage{pgfplots}
\usepackage{wrapfig}
\usepgfplotslibrary{groupplots}
\pgfplotsset{compat=1.18}
\newcolumntype{L}[1]{%
    >{\raggedright\arraybackslash}m{#1}}
\newcolumntype{C}[1]{%
    >{\centering\arraybackslash}m{#1}}
\definecolor{meanrow}{gray}{0.92}

\allowdisplaybreaks[4]

\title{MAGIC: Topology-Aware Analytic Graph Few-Shot Class-Incremental Learning}

\author{
Junlin Chen\textsuperscript{1},
Yuhan Wang\textsuperscript{1},
Xuefei Wang\textsuperscript{1},
Xiao Wang\textsuperscript{1},
Ruijie Wang\textsuperscript{1},
Jianxin Li\textsuperscript{1}\\
\textsuperscript{1}Beihang University, Beijing, China
}

\iclrfinalcopy 
\begin{document}

\maketitle

\begin{abstract}
Graph few-shot class-incremental learning (GFSCIL) requires a model to
continually recognize emerging classes from only a few labeled nodes
while preserving previously acquired knowledge. Beyond the
catastrophic forgetting inherited from conventional graph continual
learning, GFSCIL presents two distinctive challenges: extremely limited
novel-class supervision causes severe overfitting, while cross-session
edges---edges connecting newly arriving nodes with historical
nodes---alter historical propagation neighborhoods and thereby induce
representation drift. We propose MAGIC, a replay-free GFSCIL framework that combines a frozen
graph representation backbone (e.g., an intrinsically parameter-free
backbone such as SGC or a pretrained graph foundation model) with
closed-form analytic continual learning. To alleviate 
novel-session overfitting, MAGIC learns a topological prior from the base graph 
that can characterize both homophilous and heterophilous relations, and injects this prior through Potts Markov random field inference to refine supervision for novel 
classes. To mitigate representation drift, MAGIC transfers previous
predictions from the old representations of affected historical nodes
to their updated representations through drift-aware analytic
distillation. Experiments across five datasets and eight baselines demonstrate the effectiveness of MAGIC. Under the 5-shot setting, MAGIC improves Mean Accuracy and Final Accuracy by 5.48 percentage points and 9.33 percentage points on average, and reduces Performance Drop by 10.78 percentage points on average compared with the best baselines. MAGIC also shows clear advantages under the 1- and 3-shot settings, with larger gains as the number of supports increases. Moreover, MAGIC requires substantially less training time.
\end{abstract}

\section{Introduction}
\label{sec:introduction}

Graph neural networks (GNNs) have achieved remarkable success in node
classification by integrating node attributes with relational
information~\citep{kipf2017gcn,hamilton2017inductive,velickovic2018graph}.
Most conventional GNNs, however, are developed under the assumption that
the input graph and its label space remain static after training. This
assumption rarely holds in real applications, where new entities and
relations arrive continuously and the underlying graph evolves over
time. Continual graph learning (CGL) therefore aims to incorporate
newly arriving graph data while retaining knowledge acquired from
previous stages~\citep{zhou2021ergnn,liu2021twp,zhang2022cglb}.

An especially challenging yet practical setting arises when novel node
classes emerge with only a handful of labeled nodes, known as
\emph{graph few-shot class-incremental learning} (GFSCIL)
~\citep{tan2022gfcil,lu2022geometer}. In GFSCIL, a model first learns a
set of base classes from a relatively well-labeled base graph. It then
receives a sequence of incremental sessions, each introducing new nodes,
new edges, and disjoint novel classes, but only limited labeled support nodes
for each novel class. After every session, the model must classify all
classes observed so far without revisiting the complete labeled data of
previous sessions. Beyond the catastrophic forgetting problem inherited from GCIL, GFSCIL introduces two distinctive challenges.

\noindent\textbf{Challenge 1: Novel-class overfitting.}
Each novel class is represented by only a few labeled support nodes,
which provide insufficient supervision for estimating a reliable
decision boundary. The model can therefore overfit the limited supports.

\noindent\textbf{Challenge 2: Topology-induced representation drift.}
Nodes arriving in a new session may form edges with historical nodes,
thereby altering the propagation neighborhoods of historical nodes and
causing their representations to drift.

To address \noindent\textbf{Challenge 1}, existing GFSCIL methods employ
pseudo-incremental meta-training and hierarchical
attention~\citep{tan2022gfcil}, geometrically constrained prototype
learning~\citep{lu2022geometer}, structured prototype memories with
node--memory interaction~\citep{li2024mecoin}, or iterative prototype
calibration using pseudo-labeled query nodes~\citep{li2025inductive}.
However, these mechanisms operate primarily in the node-feature space
and no longer care about the graph after feature extraction. They therefore overlook the potential of graph topological patterns as a prior for alleviating novel-session overfitting.

\noindent\textbf{Challenge 2} remains largely overlooked in GFSCIL.
Geometer employs teacher--student distillation, but evaluates the
teacher on the current-session graph, which already contains newly
arrived cross-session edges that were absent when the teacher was
learned~\citep{lu2022geometer}. Mecoin freezes the GNN encoder, yet
fixed parameters do not prevent historical node representations from
drifting when cross-session edges alter their
neighborhoods~\citep{li2024mecoin}. GFCIL observes that a frozen encoder
may produce overlapping representations when novel nodes are densely
connected to base nodes, but does not model the reverse effect whereby
these connections alter historical node
representations~\citep{tan2022gfcil}.TAP instead considers an inductive setting with inaccessible or incomplete topology and therefore does not encounter the same issue~\citep{li2025inductive}.

Motivated by these observations, we propose \textbf{MAGIC} (\textbf{M}arkov
\textbf{A}nalytic \textbf{G}raph-\textbf{I}ncremental
\textbf{C}lassifier). MAGIC employs a frozen graph representation backbone
to extract node features and employs analytic learning~\citep{zhuang2022acil} to accumulate knowledge through closed-form updates, thereby mitigating
catastrophic forgetting without replay. However, as mentioned in \noindent\textbf{Challenge 2}, a fixed encoder
does not guarantee invariant historical representations, MAGIC therefore identifies the affected historical nodes and performs drift-aware analytic distillation~\citep{hinton2015distilling}: predictions produced by the previous classifier on the old graph are transferred as soft targets to their updated representations.
To alleviate overfitting in novel sessions, MAGIC further learns from
the base session a topological prior capable of characterizing
both homophilous and heterophilous relations. It then injects this prior
into the model through Potts Markov random field
inference~\citep{wang2021pmrf,wainwright2008graphical}.

Our main contributions are summarized as follows:
\begin{itemize}
    \item We propose \textbf{MAGIC}, a replay-free GFSCIL framework with analytic continual learning.
    
    \item We develop a method for estimating a topological prior and inject it through Potts Markov random field inference to alleviate novel-session overfitting.

    \item To address the representation changes caused by newly
    arriving cross-session edges, we introduce analytic
    distillation.
    
    \item Extensive experiments across five datasets and eight baselines demonstrate that MAGIC consistently outperforms existing methods under the 1-, 3-, and 5-shot settings, and achieves substantially greater training efficiency.
\end{itemize}

\section{Preliminaries}
\label{sec:preliminaries}

\noindent\textbf{Graph Few-Shot Class-Incremental Learning}
\par\nopagebreak
\noindent

The GFSCIL scenario is constructed by partitioning a complete graph
$\mathcal{G}=(\mathcal{V},\mathcal{E},\mathbf X,\mathbf y)$ into a base session and a sequence of novel sessions, where
$\mathcal{V}$ and $\mathcal{E}$ are the node and edge sets,
$\mathbf X\in\mathbb R^{|\mathcal V|\times d_x}$ is the node-attribute
matrix, and $\mathbf y$ contains the node labels. Following the GFSCIL
setting~\citep{tan2022gfcil,lu2022geometer}, the label space is
partitioned into one base-class set and $T$ mutually disjoint
novel-class sets: $\mathcal C = \mathcal C_0 \mathbin{\dot\cup} \mathcal C_1 \mathbin{\dot\cup}\cdots\mathbin{\dot\cup} \mathcal C_T$, with $\mathcal C_{\leq t} = \bigcup_{s=0}^{t}\mathcal C_s$.

Let $b(v)$ denote the session in which node $v$ arrives, with
$b(v)=t$ if $y_v\in\mathcal C_t$. At session $t$, the visible graph is
\begin{equation}
\label{eq:visible_graph}
    \mathcal G_t
    =
    (\mathcal V_t,\mathcal E_t,\mathbf X_t),
    \qquad
    \mathcal V_t
    =
    \{v\in\mathcal V:b(v)\leq t\},
    \qquad
    \mathcal E_t
    =
    \{(u,v)\in\mathcal E:
      \max(b(u),b(v))\leq t\}.
\end{equation}

In the base session, a subset $\mathcal L_c\subset\mathcal V_0$ is
labeled for every $c\in\mathcal C_0$. At each novel session $t>0$,
every class $c\in\mathcal C_t$ provides a support set
$\mathcal S_c\subset\mathcal V_t$ containing exactly $K$ labeled nodes.
The labels of all remaining nodes are unavailable. Historical labeled
nodes cannot be stored or replayed after their supervision has been
incorporated.

After session $t$, the learner produces a classifier
$f_t:\mathcal V_t\rightarrow\mathcal C_{\leq t}$. Predictions are made
over the complete set of classes encountered so far without a task
identifier.

\noindent\textbf{Frozen Graph Representation Backbones}
\par\nopagebreak
\noindent

MAGIC is compatible with a broad class of frozen graph representation
backbones. These
backbones may be intrinsically parameter-free graph propagation
operators, such as SGC, Simple Spectral Graph Convolution
(S\textsuperscript{2}GC), personalized PageRank diffusion and Heat Kernel
~\citep{wu2019sgc,zhu2021ssgc,gasteiger2019ppnp,xu2019graphheat}, or pretrained graph foundation models used with frozen parameters, such as AnyGraph and RGVT~\citep{xia2026anygraph,lee2026rgvt}. 

\section{Methodology}
\label{sec:methodology}

\begin{figure}[t]
    \centering
    \includegraphics[width=0.84\linewidth]{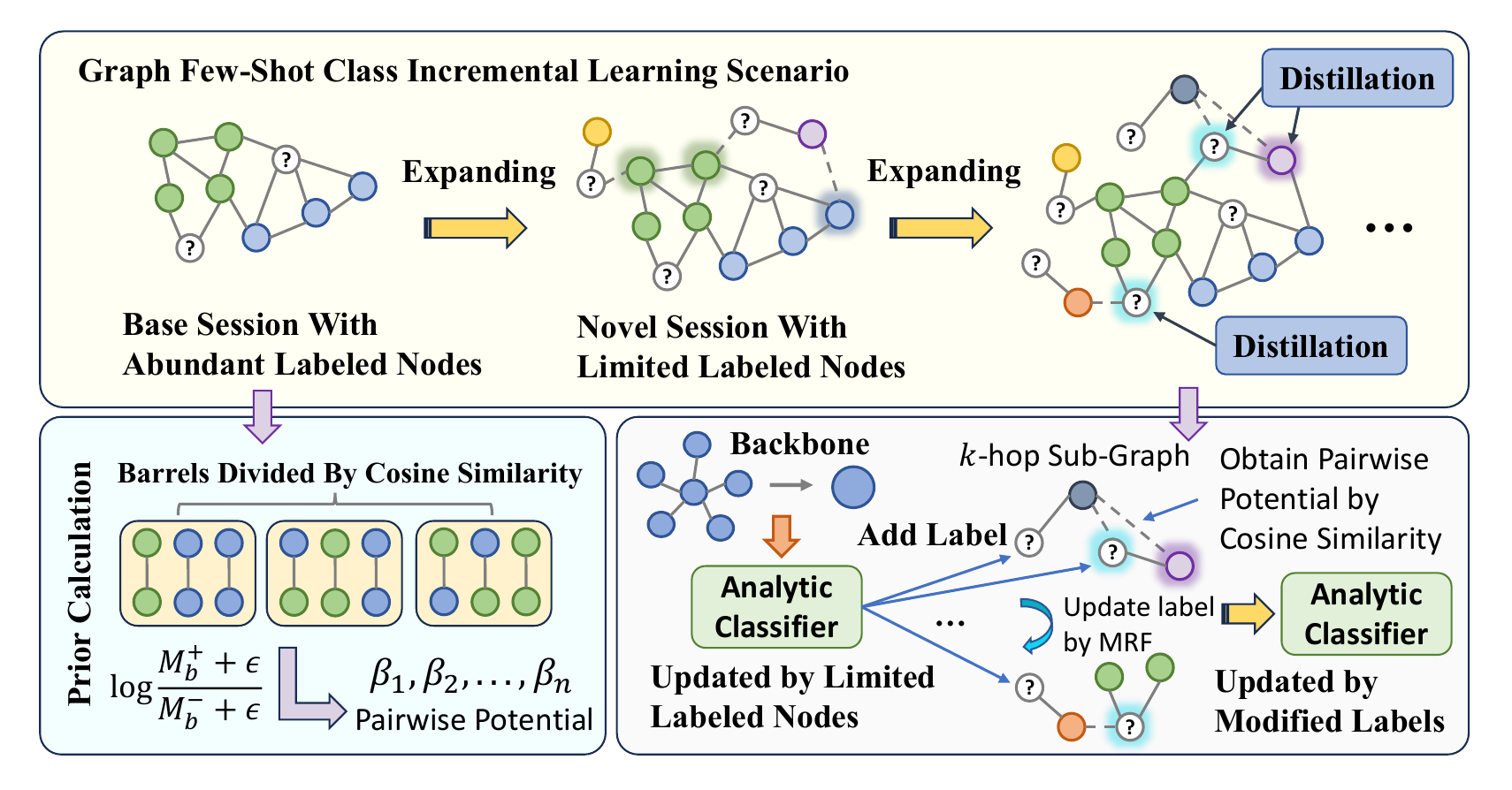}
    \caption{On the base stage, MAGIC learns a topological prior. In novel sessions, limited labeled supports first update an initial classifier, while drift-aware distillation protects historical nodes affected by cross-session edges. Finally, MAGIC extracts a subgraph around novel supports and performs Potts MRF inference with the topological prior to adjust the pseudo-labels predicted by the initial classifier, and subsequently updates the classifier using these adjusted labels.}
    \label{fig:overall_framework}
\end{figure}

MAGIC is motivated by the following design rationale. It combines fixed graph encoder 
with analytic continual learning to preserve accumulated knowledge
through closed-form updates and mitigate catastrophic forgetting without
replay. Since cross-session edges can still alter historical node
representations, MAGIC employs analytic distillation to transfer
previous predictions to their updated representations. To alleviate
novel-session overfitting, MAGIC estimates the relationship between the
cosine similarity of connected node representations and their label
agreement. As this relation is independent of specific class identities,
it is more likely to generalize to novel classes. MAGIC then injects the
resulting topological prior through MRF inference. The overall framework is illustrated in Figure~\ref{fig:overall_framework}.

\subsection{Base Stage: Analytic Learning and Relational Prior}
\label{sec:base_stage}

Let $\mathcal G_0=(\mathcal V_0,\mathcal E_0)$ be the base graph,
$\mathcal C_0$ the set of base classes, and $\mathbf h_i\in\mathbb R^d$
the representation of node $i$. For class $c$, denote its
labeled node set by $\mathcal L_c$ and its size by $n_c$.

\paragraph{Class-balanced analytic classifier.}
Directly assigning the same weight to every labeled node would make
large base classes dominate the analytic solution. Moreover, each novel
class subsequently contributes unit support mass and can receive at
most $\rho$ additional pseudo-supervision mass (Further details are provided in Section~\ref{sec:novel_stage}.). MAGIC therefore assigns
each base class total hard-label mass $1+\rho$, establishing the same
maximum direct-supervision budget across base and novel classes.
Accordingly, every labeled base node of class $c$ receives weight
$(1+\rho)/n_c$. The initial sufficient statistics and classifier are
\begin{equation}
\label{eq:base_analytic}
    \mathbf R_0
    =
    \sum_{c\in\mathcal C_0}
    \sum_{i\in\mathcal L_c}
    \frac{1+\rho}{n_c}\mathbf h_i\mathbf h_i^\top,
    \qquad
    \mathbf C_0
    =
    \sum_{c\in\mathcal C_0}
    \sum_{i\in\mathcal L_c}
    \frac{1+\rho}{n_c}\mathbf h_i\mathbf e_c^\top,
    \qquad
    \mathbf W_0
    =
    (\mathbf R_0+\lambda\mathbf I)^{-1}\mathbf C_0,
\end{equation}
where $\mathbf e_c$ is the one-hot vector of class $c$,
$\lambda>0$ is the ridge coefficient, and $\rho\geq0$ is the
pseudo-supervision budget defined in
Equation~\ref{eq:pseudo_balance}. The sufficient statistics
$\mathbf R_0$ and $\mathbf C_0$ summarize the base supervision, so the
original base nodes need not be replayed in subsequent sessions.

\paragraph{Class-balanced relation statistics.}
MAGIC further learns how a connected node pair's label agreement varies
with the cosine similarity of its representations. We collect edges
whose two endpoints are labeled base nodes and partition their cosine
similarities into $B$ quantile intervals
$\{\mathcal I_1,\ldots,\mathcal I_B\}$. Quantile partitioning prevents
the relation statistics from being concentrated in a small number of
similarity intervals.

A naive edge count would again be biased toward large classes. For a
same-class edge $(i,j)$ from class $c$, we therefore assign the
opportunity-normalized weight
$w_{ij}^{+}=1/(|\mathcal C_0|\binom{n_c}{2})$. For an edge connecting
classes $c\neq c'$, we use
$w_{ij}^{-}=1/(\binom{|\mathcal C_0|}{2}n_cn_{c'})$. These weights are
equivalent to first sampling a class or class pair uniformly and then
sampling nodes within the selected classes.

For similarity interval $\mathcal I_b$, the balanced same-class and
different-class edge masses are
\begin{equation}
M_b^{+}=\sum_{(i,j)\in\mathcal E_0}w_{ij}^{+}\mathbf 1\{y_i=y_j\}
\mathbf 1\{\mathbf h_i^\top\mathbf h_j\in\mathcal I_b\},
\quad
M_b^{-}=\sum_{(i,j)\in\mathcal E_0}w_{ij}^{-}\mathbf 1\{y_i\neq y_j\}
\mathbf 1\{\mathbf h_i^\top\mathbf h_j\in\mathcal I_b\}.
\end{equation}
The signed relation coefficient is then
\begin{equation}
\label{eq:relation_prior}
    \beta_b
    =
    \operatorname{clip}
    \left(
        \log\frac{M_b^{+}+\epsilon}{M_b^{-}+\epsilon},
        -\beta_{\max},
        \beta_{\max}
    \right),
\end{equation}
where $\epsilon>0$ stabilizes sparsely populated intervals.

A positive $\beta_b$ favors equal labels, whereas a negative value
favors different labels. MAGIC therefore does not impose a homophily
assumption and can adapt to homophilous, heterophilous, or mixed graph
relations. Moreover, $\beta_b$ depends only on label agreement rather
than class identity. It is consequently invariant to class permutation and is more likely to
transfer from base classes to unseen classes.

\subsection{Novel Stage: Analytic Learning and Topological Prior Injection}
\label{sec:novel_stage}

\noindent\textbf{Step I: Naive Classifier with Drift Distillation}
\par\nopagebreak
\noindent

MAGIC first incorporates the labeled novel supports into the
accumulated analytic statistics. Since each novel class provides
exactly $K$ supports, every support is assigned mass $1/K$. Before this update, zero columns for the newly introduced classes are appended to $\mathbf C_{t-1}$.

Although the graph backbone is either parameter-free or used with
frozen parameters, newly arriving
cross-session edges may change the propagation neighborhoods and thus
the representations of historical nodes. Let
$\mathbf h_{t-1,i}$ and $\mathbf h_{t,i}$ denote the previous and current
representations of an affected historical node $i$, respectively. MAGIC
measures its topology-induced drift as
$\delta_i=\|\mathbf h_{t,i}-\mathbf h_{t-1,i}\|_2$ and assigns the
normalized distillation mass
\begin{equation}
a_i^{D}
=
\kappa |\mathcal C_t^{\mathrm{new}}|
\frac{\delta_i}
{\sum_{j\in\mathcal A_t}\delta_j},
\qquad i\in\mathcal A_t,
\label{eq:distill_mass}
\end{equation}
where $\mathcal A_t$ is the set of affected historical nodes and $\kappa$ determines the total
distillation budget. The teacher target is obtained from the previous
classifier and temperature-adjusted as
\begin{equation}
\mathbf q_i^{D}
=
\operatorname{Pad}_{|\mathcal C_{\le t}|}
\left[
\operatorname{Temp}_{T_D}
\left(
\Pi_{\Delta}
\left(\mathbf h_{t-1,i}^{\top}\mathbf W_{t-1}\right)
\right)
\right],
\quad
\operatorname{Temp}_{T_D}(\mathbf p)
=
\frac{\mathbf p^{1/T_D}}
{\sum_c p_c^{1/T_D}} .
\label{eq:distill_target}
\end{equation}
where $\Pi_{\Delta}$ projects scores onto the probability simplex and
$\operatorname{Pad}$ appends zeros for newly introduced classes. The
resulting soft class mass $\boldsymbol{\alpha}_i^D=a_i^D\mathbf q_i^D$
is imposed on the current representation by updating the analytic
sufficient statistics:
\begin{equation}
\mathbf R_t \leftarrow \mathbf R_t + \sum_{i\in\mathcal A_t} a_i^D\mathbf h_{t,i}\mathbf h_{t,i}^{\top},
\quad
\mathbf C_t \leftarrow \mathbf C_t + \sum_{i\in\mathcal A_t} \mathbf h_{t,i}(\boldsymbol{\alpha}_i^D)^{\top}.
\label{eq:distill_statistics}
\end{equation}
Consequently, nodes undergoing larger topology-induced drift receive
stronger protection while historical predictions are transferred
without introducing artificial supervision for novel classes.

The influence of cross-session edges on analytic learning is summarized
by the following theorem.

\begin{theorem}[Topology-induced perturbation]
\label{thm:topology_drift}
Let $\Phi_{\boldsymbol\theta^\star}$ be a fixed graph encoder, and let
$\mathbf H^{-}$ and
$\mathbf H^{+}=\mathbf H^{-}+\boldsymbol{\Delta}$ be the representations
of a set of historical nodes before and after graph evolution.
Given supervision matrix $\boldsymbol{\Gamma}$ and
$\mathbf D=\operatorname{diag}(\boldsymbol{\Gamma}\mathbf 1)$, their
analytic statistics satisfy
\begin{align}
    \mathbf R^{+}-\mathbf R^{-}
    &=
    (\mathbf H^{-})^\top\mathbf D\boldsymbol{\Delta}
    +
    \boldsymbol{\Delta}^\top\mathbf D\mathbf H^{-}
    +
    \boldsymbol{\Delta}^\top\mathbf D\boldsymbol{\Delta},
    \nonumber\\
    \mathbf C^{+}-\mathbf C^{-}
    &=
    \boldsymbol{\Delta}^\top\boldsymbol{\Gamma}.
\end{align}
\end{theorem}

Theorem~\ref{thm:topology_drift} shows that a fixed graph encoder does
not imply fixed analytic statistics on a dynamic graph, motivating the
proposed drift-aware distillation. Its proof is provided in
Appendix~\ref{app:theorem_proofs}.

After adding the class-balanced novel supports and distillation targets
to the accumulated statistics, MAGIC obtains the naive classifier
$\mathbf W_t^{(0)}$ through the same closed-form solution as
Equation~\ref{eq:base_analytic}.

\noindent\textbf{Step II: MRF Refinement and Analytic Reinjection}
\par\nopagebreak
\noindent

Starting from all novel-session supports, MAGIC extracts their $k$-hop induced subgraph from the
currently visible graph. The subgraph can contain both historical and
current-session nodes because the stage identity of an unlabeled node
is not assumed to be known.

For every node $i$ in this subgraph, the naive classifier produces a
probability $\mathbf p_{0,i}$ over all classes observed so far, it is obtained using the same simplex projection as in
Equation~\eqref{eq:distill_target}, but without temperature adjustment. This
serves as the unary evidence. For a subgraph edge $(i,j)$, MAGIC selects
the base relation coefficient $\beta_{b(i,j)}$ according to the current
cosine similarity and constructs the degree-normalized coupling
$J_{ij}=\beta_{b(i,j)}/\sqrt{d_id_j}$. The resulting Potts MRF is
\begin{equation}
\label{eq:mrf_distribution}
    p(\mathbf y)
    \propto
    \prod_i p_{0,i}(y_i)^{\tau_u}
    \prod_{(i,j)\in\mathcal E_t^{\mathrm{sub}}}
    \exp\left(
        \eta J_{ij}\mathbf 1\{y_i=y_j\}
    \right),
\end{equation}
where support nodes are clamped to their ground-truth labels. Degree
normalization prevents hubs from accumulating excessive relational
evidence.

MAGIC approximates the node marginals using damped mean-field
inference:
\begin{equation}
\label{eq:mean_field}
    \widehat{\mathbf q}_i^{(\ell+1)}
    =
    \operatorname{softmax}
    \left(
        \tau_u\log\mathbf p_{0,i}
        +
        \eta\sum_{j\in\mathcal N_i}J_{ij}
        \mathbf q_j^{(\ell)}
    \right),
    \qquad
    \mathbf q_i^{(\ell+1)}
    =
    (1-\alpha)\mathbf q_i^{(\ell)}
    +
    \alpha\widehat{\mathbf q}_i^{(\ell+1)}.
\end{equation}
The support nodes are reset to their one-hot labels after every
iteration.

\begin{theorem}[Convergence of mean-field inference]
\label{thm:mean_field_convergence}
Let $\mathbf J_{\mathrm{FF}}$ be the coupling matrix among free,
unclamped nodes. If
$\eta\|\mathbf J_{\mathrm{FF}}\|_2/2<1$, the damped mean-field mapping is
a contraction under the Frobenius norm and converges to a unique fixed
point. Its contraction factor is at most
$1-\alpha+\alpha\eta\|\mathbf J_{\mathrm{FF}}\|_2/2<1$.
\end{theorem}

The proof is deferred to Appendix~\ref{app:theorem_proofs}. The complete
probabilistic derivation is provided in
Appendix~\ref{app:probabilistic_interpretation}.

\paragraph{Novel-class posterior projection.}
The support-centered subgraph may contain both historical and
current-session nodes. Labeled novel supports provide reliable
class-specific anchors whose supervision can propagate to surrounding
nodes through the MRF, whereas no labeled historical nodes are
available to provide corresponding old-class anchors. MAGIC therefore
uses the inferred posterior to construct additional supervision for
the current novel classes, rather than reinforcing historical-class
predictions without labeled support.

Specifically, mean-field inference produces a calibrated posterior
$\mathbf q_i$ over all classes observed up to session $t$. For every
unlabeled candidate node, MAGIC retains its novel-class components
without renormalization:
\begin{equation}
\label{eq:novel_projection}
    \widetilde{\gamma}_{i,c}
    =
    \begin{cases}
        q_i(c), & c\in\mathcal C_t^{\mathrm{new}},\\
        0,      & c\notin\mathcal C_t^{\mathrm{new}}.
    \end{cases}
\end{equation}
This projection does not preselect or discard candidate nodes. Instead,
the retained mass
$\sum_{c\in\mathcal C_t^{\mathrm{new}}}q_i(c)$ directly determines how
strongly node $i$ contributes to novel-class learning: nodes whose
posterior is concentrated on historical classes naturally provide
little novel-class supervision.

\paragraph{Conservative class balancing.}
The projected posteriors may assign substantially different total
masses to different novel classes. Let
$M_c=\sum_i\widetilde{\gamma}_{i,c}$ denote the available posterior mass
for novel class $c$. MAGIC applies
\begin{equation}
\label{eq:pseudo_balance}
    \gamma_{i,c}^{\mathrm{mrf}}
    =
    \min\left(1,\frac{\rho}{M_c+\epsilon}\right)
    \widetilde{\gamma}_{i,c},
    \qquad c\in\mathcal C_t^{\mathrm{new}}.
\end{equation}
This limits each novel class to at most $\rho$ pseudo-supervision mass
while preserving the relative posterior allocation among candidate
nodes. When the available evidence for a class is weaker than $\rho$,
it remains unchanged rather than being artificially amplified.

Finally, let $\mathcal U_t$ denote the unlabeled candidate nodes and
$m_i^{\mathrm{mrf}}
=\|\boldsymbol{\gamma}_i^{\mathrm{mrf}}\|_1$ their supervision masses.
The MRF-refined supervision is incorporated through
\begin{equation}
\mathbf R_t
\leftarrow
\mathbf R_t+
\sum_{i\in\mathcal U_t}
m_i^{\mathrm{mrf}}\mathbf h_i\mathbf h_i^\top,
\qquad
\mathbf C_t
\leftarrow
\mathbf C_t+
\sum_{i\in\mathcal U_t}
\mathbf h_i
(\boldsymbol{\gamma}_i^{\mathrm{mrf}})^\top .
\end{equation}
MAGIC then obtains
$\mathbf W_t=(\mathbf R_t+\lambda\mathbf I)^{-1}\mathbf C_t$.
Thus, hard labels, distillation targets, and MRF-refined soft
supervision are incorporated through the same analytic statistics.

\section{Experiments}
\label{sec:experiments}

\subsection{Setup}
\label{sec:experimental_setup}
\noindent\textbf{Datasets.}
We conduct experiments on five node-classification benchmarks:
CoraFull~\citep{bojchevski2018deep}, Coauthor CS and Amazon
Computers~\citep{shchur2018pitfalls}, ogbn-arxiv~\citep{hu2020ogb},
and WikiCS~\citep{mernyei2020wikics}.

\noindent\textbf{Baselines.}
We compare MAGIC with general and graph continual-learning methods,
including LwF~\citep{li2016learning}, EWC~\citep{kirkpatrick2017overcoming},
MAS~\citep{aljundi2018memory}, and TWP~\citep{liu2021twp}. We also
compare against GFSCIL methods, including GFCIL~\citep{tan2022gfcil},
Geometer~\citep{lu2022geometer}, Mecoin~\citep{li2024mecoin}, and
TAP~\citep{li2025inductive}.

\noindent\textbf{Evaluation Metrics.}
We report three evaluation metrics. \textbf{MA (Mean Accuracy)} is the average
overall accuracy across all sessions, and \textbf{FA (Final Accuracy)} is the
overall accuracy after the final session. \textbf{PD} denotes the
performance drop from the base session to the final session, with a
smaller value indicating better retention. More details about datasets, baselines and backbones, and hyperparameter sensitivity analysis are shown in Appendix~\ref{sec:further_experiments}

\subsection{Main Results}

Table~\ref{tab:main_5shot_all} reports the 5-shot results on all five
datasets. 

\begin{table}[t]
    \caption{Main results under the 5-shot setting. MAGIC Backbone: SGC. MAGIC improves Mean Accuracy and Final Accuracy by 5.48 percentage points and 9.33 percentage points on average, and reduces Performance Drop by 10.78 percentage points on average compared with the best baselines.}
    \label{tab:main_5shot_all}
    \centering
    \scriptsize
    \setlength{\tabcolsep}{1.2pt}
    \renewcommand{\arraystretch}{1.08}
    \begin{tabular}{L{1.70cm}*{15}{C{0.70cm}}}
        \toprule
        & \multicolumn{3}{c}{CoraFull}
        & \multicolumn{3}{c}{ogbn-arxiv}
        & \multicolumn{3}{c}{Coauthor CS}
        & \multicolumn{3}{c}{Amazon Computers}
        & \multicolumn{3}{c}{WikiCS} \\
        \cmidrule(lr){2-4}
        \cmidrule(lr){5-7}
        \cmidrule(lr){8-10}
        \cmidrule(lr){11-13}
        \cmidrule(lr){14-16}
        Method
        & MA & FA & PD
        & MA & FA & PD
        & MA & FA & PD
        & MA & FA & PD
        & MA & FA & PD \\
        \midrule

        \rowcolor{meanrow}
        LwF
        & 16.06 & 3.82 & 79.86
        & 18.16 & 4.65 & 82.03
        & 35.83 & 17.16 & 79.73
        & 35.05 & 13.55 & 82.07
        & 35.83 & 12.90 & 79.78 \\
        LwF (std)
        & $\pm$1.24 & $\pm$1.44 & $\pm$2.35
        & $\pm$0.97 & $\pm$5.80 & $\pm$6.57
        & $\pm$3.44 & $\pm$5.45 & $\pm$5.73
        & $\pm$1.73 & $\pm$5.61 & $\pm$4.53
        & $\pm$2.33 & $\pm$3.96 & $\pm$4.33 \\

        \rowcolor{meanrow}
        EWC
        & 16.14 & 3.86 & 79.82
        & 15.87 & 6.07 & 80.52
        & 36.38 & 17.32 & 79.58
        & 35.13 & 13.96 & 81.54
        & 35.89 & 12.86 & 79.82 \\
        EWC (std)
        & $\pm$1.31 & $\pm$1.39 & $\pm$2.30
        & $\pm$2.23 & $\pm$6.70 & $\pm$7.19
        & $\pm$3.85 & $\pm$5.54 & $\pm$5.78
        & $\pm$1.74 & $\pm$5.92 & $\pm$4.90
        & $\pm$2.49 & $\pm$3.86 & $\pm$4.27 \\

        \rowcolor{meanrow}
        MAS
        & 16.31 & 6.07 & 77.61
        & 34.40 & 27.37 & 59.25
        & 45.91 & 20.77 & 76.12
        & 34.98 & 13.35 & 82.17
        & 35.24 & 12.42 & 80.26 \\
        MAS (std)
        & $\pm$1.21 & $\pm$1.93 & $\pm$2.13
        & $\pm$5.04 & $\pm$10.00 & $\pm$7.75
        & $\pm$5.15 & $\pm$5.52 & $\pm$5.51
        & $\pm$1.88 & $\pm$5.41 & $\pm$4.19
        & $\pm$2.04 & $\pm$4.05 & $\pm$4.50 \\

        \rowcolor{meanrow}
        TWP
        & 17.48 & 5.05 & 78.21
        & 32.36 & 17.24 & 68.38
        & 39.70 & 15.93 & 80.97
        & 34.76 & 13.37 & 82.45
        & 35.80 & 13.24 & 78.94 \\
        TWP (std)
        & $\pm$1.49 & $\pm$1.67 & $\pm$3.38
        & $\pm$8.83 & $\pm$5.84 & $\pm$5.58
        & $\pm$5.54 & $\pm$3.59 & $\pm$3.78
        & $\pm$1.50 & $\pm$5.54 & $\pm$4.14
        & $\pm$1.82 & $\pm$3.71 & $\pm$3.52 \\

        \rowcolor{meanrow}
        GFCIL
        & 54.23 & 41.33 & 39.81
        & 56.77 & 40.78 & 42.02
        & 78.02 & 71.29 & 24.95
        & \textbf{86.72} & 80.01 & 12.61
        & 76.63 & 67.18 & 23.74 \\
        GFCIL (std)
        & $\pm$1.64 & $\pm$2.28 & $\pm$5.31
        & $\pm$4.75 & $\pm$3.34 & $\pm$3.56
        & $\pm$3.98 & $\pm$3.77 & $\pm$5.35
        & \textbf{$\pm$2.79} & $\pm$2.20 & $\pm$4.13
        & $\pm$2.77 & $\pm$2.85 & $\pm$3.71 \\

        \rowcolor{meanrow}
        Geometer
        & 50.10 & 34.05 & 46.27
        & 36.29 & 14.55 & 55.68
        & 81.86 & 71.39 & 25.27
        & 78.08 & 63.65 & 26.97
        & 70.16 & 51.41 & 37.50 \\
        Geometer (std)
        & $\pm$1.74 & $\pm$2.76 & $\pm$4.73
        & $\pm$2.21 & $\pm$3.51 & $\pm$6.66
        & $\pm$3.49 & $\pm$4.20 & $\pm$3.21
        & $\pm$4.26 & $\pm$4.16 & $\pm$4.62
        & $\pm$4.74 & $\pm$8.11 & $\pm$8.20 \\

        \rowcolor{meanrow}
        Mecoin
        & 47.57 & 32.60 & 44.13
        & 46.20 & 31.08 & 46.18
        & 71.81 & 54.16 & 42.07
        & 70.44 & 56.31 & 32.13
        & 65.17 & 50.12 & 38.62 \\
        Mecoin (std)
        & $\pm$1.60 & $\pm$1.63 & $\pm$4.54
        & $\pm$3.89 & $\pm$0.89 & $\pm$4.90
        & $\pm$4.00 & $\pm$1.86 & $\pm$2.33
        & $\pm$2.00 & $\pm$3.86 & $\pm$3.68
        & $\pm$3.97 & $\pm$4.55 & $\pm$4.65 \\

        \rowcolor{meanrow}
        TAP
        & 59.27 & 45.06 & 38.48
        & 57.97 & 42.69 & 42.37
        & 79.81 & 67.83 & 28.76
        & 83.76 & 74.06 & 21.10
        & 76.24 & 63.66 & 28.16 \\
        TAP (std)
        & $\pm$1.35 & $\pm$1.96 & $\pm$4.04
        & $\pm$5.09 & $\pm$1.23 & $\pm$3.74
        & $\pm$1.87 & $\pm$3.56 & $\pm$4.04
        & $\pm$5.27 & $\pm$4.90 & $\pm$4.03
        & $\pm$3.27 & $\pm$5.56 & $\pm$6.78 \\

        \rowcolor{meanrow}
        \textbf{MAGIC}
        & \textbf{65.45} & \textbf{56.02} & \textbf{24.19}
        & \textbf{63.50} & \textbf{49.81} & \textbf{34.37}
        & \textbf{94.03} & \textbf{91.45} & \textbf{5.54}
        & 86.54 & \textbf{81.04} & \textbf{9.70}
        & \textbf{80.33} & \textbf{74.64} & \textbf{14.09} \\
        \textbf{MAGIC (std)}
        & \textbf{$\pm$0.95} & \textbf{$\pm$1.05} & \textbf{$\pm$3.65}
        & \textbf{$\pm$3.92} & \textbf{$\pm$2.35} & \textbf{$\pm$2.55}
        & \textbf{$\pm$0.79} & \textbf{$\pm$0.28} & \textbf{$\pm$1.34}
        & $\pm$1.68 & \textbf{$\pm$2.88} & \textbf{$\pm$5.06}
        & \textbf{$\pm$1.74} & \textbf{$\pm$2.14} & \textbf{$\pm$3.01} \\
        \bottomrule
    \end{tabular}
\end{table}

MAGIC achieves the best MA, FA, and PD on CoraFull, ogbn-arxiv,
Coauthor CS, and WikiCS. On Amazon Computers, it obtains the strongest
FA and PD while remaining competitive in MA. Conventional continual
learning methods deteriorate substantially because preserving previous
parameters alone does not provide sufficient supervision for emerging
classes, while GFSCIL-specific methods perform considerably better.
Across graphs with different scales and structural characteristics,
MAGIC consistently outperforms existing methods, demonstrating the
effectiveness and robustness of the proposed framework.

\begin{figure}[t]
    \centering
    \begin{minipage}[t]{0.49\linewidth}
        \vspace{0pt}
        \centering
        \includegraphics[width=\linewidth]
        {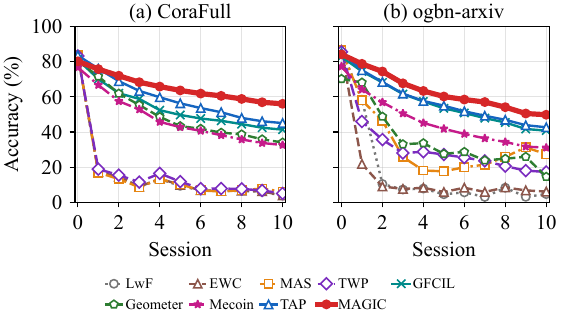}
        \caption{Overall accuracy across the base session and ten
        incremental sessions under the 5-shot setting. MAGIC Backbone: SGC}
        \label{fig:session_accuracy}
    \end{minipage}
    \hfill
    \begin{minipage}[t]{0.49\linewidth}
        \vspace{0pt}
        \centering
        \includegraphics[width=\linewidth]
        {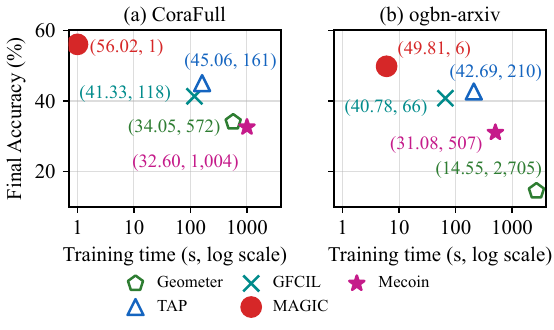}
        \caption{Final accuracy versus total training time under the
        5-shot setting. MAGIC Backbone: SGC}
        \label{fig:training_efficiency}
    \end{minipage}
\end{figure}

Figure~\ref{fig:session_accuracy} presents the session-wise accuracy
trajectories on CoraFull and ogbn-arxiv. MAGIC exhibits the most stable
trajectory on both datasets, and its advantage becomes increasingly
clear as novel classes accumulate. This indicate that MAGIC can more effectively acquire knowledge from
sparsely supervised novel sessions while preserving previously acquired
knowledge throughout the incremental process.

\begin{figure}[t]
    \centering
    \includegraphics[width=0.9\linewidth]
    {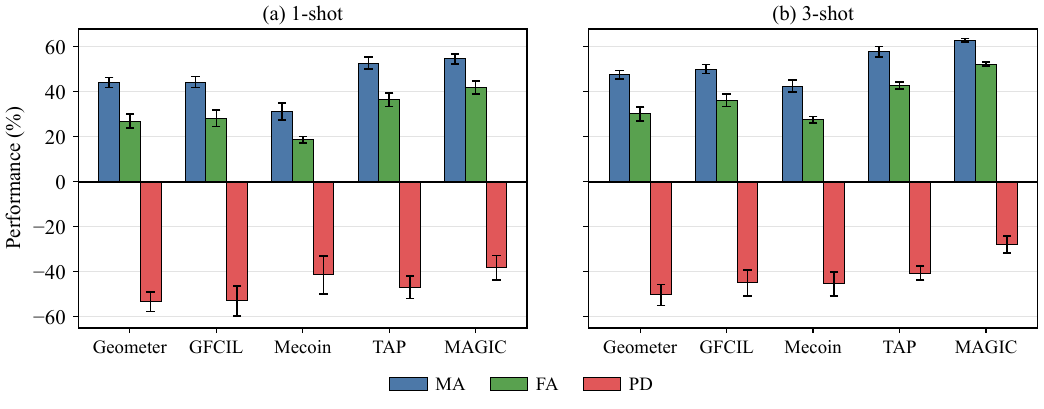}
    \caption{Main results on CoraFull under the 1- and 3-shot settings. MAGIC Backbone: SGC}
    \label{fig:main_1shot_3shot}
\end{figure}

Figure~\ref{fig:main_1shot_3shot} summarizes the 1- and 3-shot results
on CoraFull. MAGIC achieves the strongest MA and FA and the smallest PD
under both supervision levels. Taken together with the 5-shot results
in Table~\ref{tab:main_5shot_all}, the advantage becomes more
pronounced as additional supports are provided. Full results on 1- and 3-shot settings are provided in Appendix~\ref{app:additional_main_results}.

\subsection{Training Efficiency}

We compare the total training time and final accuracy of GFSCIL methods
on CoraFull and ogbn-arxiv. The results are shown in Figure~\ref{fig:training_efficiency}.

MAGIC achieves the strongest final accuracy while requiring
substantially less training time than competing GFSCIL methods on both
datasets, demonstrating a favorable balance between predictive
performance and computational efficiency.

\subsection{Effect of Frozen Graph Backbones}

We evaluate MAGIC with five frozen graph representation backbones under
the 5-shot setting. All other components and experimental protocols
remain unchanged.

\begin{table}[t]
    \caption{Results (\%) with different frozen graph backbones under
    the 5-shot setting.  RGVT is excluded from ogbn-arxiv because its released checkpoint was supervised-pretrained on that dataset.}
    \label{tab:backbone_comparison}
    \centering
    \scriptsize
    \setlength{\tabcolsep}{1.2pt}
    \renewcommand{\arraystretch}{1.08}
    \begin{tabular}{L{1.70cm}*{15}{C{0.70cm}}}
        \toprule
        & \multicolumn{3}{c}{CoraFull}
        & \multicolumn{3}{c}{ogbn-arxiv}
        & \multicolumn{3}{c}{Coauthor CS}
        & \multicolumn{3}{c}{Amazon Computers}
        & \multicolumn{3}{c}{WikiCS} \\
        \cmidrule(lr){2-4}
        \cmidrule(lr){5-7}
        \cmidrule(lr){8-10}
        \cmidrule(lr){11-13}
        \cmidrule(lr){14-16}
        Backbone
        & MA & FA & PD
        & MA & FA & PD
        & MA & FA & PD
        & MA & FA & PD
        & MA & FA & PD \\
        \midrule

        \rowcolor{meanrow}
        SGC
        & 65.45 & 56.02 & 24.19
        & 63.50 & 49.81 & 34.37
        & 94.03 & 91.45 & 5.54
        & 86.54 & 81.04 & \textbf{9.70}
        & 80.33 & 74.64 & \textbf{14.09} \\
        SGC (std)
        & $\pm$0.95 & $\pm$1.05 & $\pm$3.65
        & $\pm$3.92 & $\pm$2.35 & $\pm$2.55
        & $\pm$0.79 & $\pm$0.28 & $\pm$1.34
        & $\pm$1.68 & $\pm$2.88 & \textbf{$\pm$5.06}
        & $\pm$1.74 & $\pm$2.14 & \textbf{$\pm$3.01} \\

        \rowcolor{meanrow}
        S$^2$GC
        & 64.42 & 54.74 & 25.48
        & 61.23 & 47.05 & 36.43
        & 94.25 & 91.77 & 5.43
        & \textbf{87.26} & \textbf{81.69} & 9.76
        & \textbf{80.61} & \textbf{74.86} & 14.20 \\
        S$^2$GC (std)
        & $\pm$0.57 & $\pm$1.20 & $\pm$4.24
        & $\pm$4.01 & $\pm$2.39 & $\pm$2.66
        & $\pm$0.78 & $\pm$0.48 & $\pm$1.40
        & \textbf{$\pm$1.58} & \textbf{$\pm$2.72} & $\pm$4.79
        & \textbf{$\pm$1.55} & \textbf{$\pm$2.09} & $\pm$3.34 \\

        \rowcolor{meanrow}
        PPR
        & \textbf{66.92} & \textbf{57.62} & \textbf{22.84}
        & \textbf{65.40} & \textbf{51.34} & \textbf{34.13}
        & 94.26 & 91.90 & 5.02
        & 85.89 & 79.96 & 10.58
        & 80.12 & 73.90 & 14.98 \\
        PPR (std)
        & \textbf{$\pm$1.08} & \textbf{$\pm$0.51}
        & \textbf{$\pm$2.84}
        & \textbf{$\pm$4.13} & \textbf{$\pm$2.14}
        & \textbf{$\pm$2.52}
        & $\pm$0.63 & $\pm$0.24 & $\pm$1.42
        & $\pm$2.08 & $\pm$2.49 & $\pm$4.90
        & $\pm$1.98 & $\pm$2.61 & $\pm$3.58 \\

        \rowcolor{meanrow}
        Heat Kernel
        & 57.96 & 46.71 & 31.06
        & 53.99 & 39.30 & 40.32
        & 93.81 & 90.91 & 6.62
        & 85.56 & 78.55 & 12.66
        & 79.48 & 73.49 & 14.80 \\
        Heat Kernel (std)
        & $\pm$0.72 & $\pm$1.27 & $\pm$4.44
        & $\pm$4.41 & $\pm$3.29 & $\pm$3.41
        & $\pm$0.90 & $\pm$0.96 & $\pm$1.66
        & $\pm$1.53 & $\pm$2.72 & $\pm$4.66
        & $\pm$1.53 & $\pm$2.06 & $\pm$2.95 \\

        \rowcolor{meanrow}
        RGVT
        & 62.98 & 52.74 & 26.97
        & -- & -- & --
        & \textbf{94.57} & \textbf{92.19} & \textbf{4.99}
        & 85.83 & 79.34 & 11.73
        & 80.09 & 74.26 & 14.54 \\
        RGVT (std)
        & $\pm$0.43 & $\pm$1.13 & $\pm$4.20
        & -- & -- & --
        & \textbf{$\pm$0.53} & \textbf{$\pm$0.43}
        & \textbf{$\pm$1.54}
        & $\pm$1.97 & $\pm$2.56 & $\pm$4.50
        & $\pm$1.62 & $\pm$2.32 & $\pm$3.40 \\
        \bottomrule
    \end{tabular}
\end{table}

Table~\ref{tab:backbone_comparison} shows that MAGIC is compatible with
both parameter-free propagation operators and a frozen pretrained graph
encoder. No single backbone dominates every graph: PPR performs best on
the two citation networks, RGVT is strongest on Coauthor CS, and
S$^2$GC achieves the highest accuracy on Amazon Computers and WikiCS,
whereas SGC provides the smallest PD. These results demonstrate that
MAGIC can benefit from backbone choices adapted to different graph
characteristics.

\subsection{Ablation Study}

\begin{wraptable}{r}{0.49\textwidth}
    \centering
    \caption{Ablation results (\%) on ogbn-arxiv under the 5-shot
    setting using SGC.}
    \label{tab:ablation}
    \scriptsize
    \setlength{\tabcolsep}{2.5pt}
    \renewcommand{\arraystretch}{1.08}
    \begin{tabular}{L{1.50cm}*{3}{C{1.50cm}}}
        \toprule
        Variant & Mean Acc. & Final Acc. & PD \\
        \midrule
        w/o Prior
        & 53.69$\pm$7.74
        & 37.29$\pm$7.28
        & 46.89$\pm$6.72 \\
        w/o Distill
        & 61.88$\pm$3.61
        & 48.02$\pm$1.69
        & 36.16$\pm$4.61 \\
        MAGIC
        & \textbf{63.50$\pm$3.92}
        & \textbf{49.81$\pm$2.35}
        & \textbf{34.37$\pm$2.55} \\
        \bottomrule
    \end{tabular}
\end{wraptable}

We ablate the two main components of MAGIC on ogbn-arxiv under the
5-shot setting. \textit{w/o Prior} removes MRF-based topology-prior
injection, and \textit{w/o Distill} disables drift-aware analytic
distillation. Results are shown in Table~\ref{tab:ablation}.

Removing the topology prior causes the largest drop, confirming that the topological prior provides complementary structural
supervision for novel classes and thereby mitigates overfitting to the
limited labeled supports. Disabling distillation also degrades overall accuracy and
increases PD, showing that compensating for topology-induced
representation drift improves historical retention. The complete model achieves the strongest results on all three metrics.

\subsection{Novel Session Performance}

\begin{wrapfigure}{l}{0.49\textwidth}
    \centering
    \includegraphics[width=\linewidth]
    {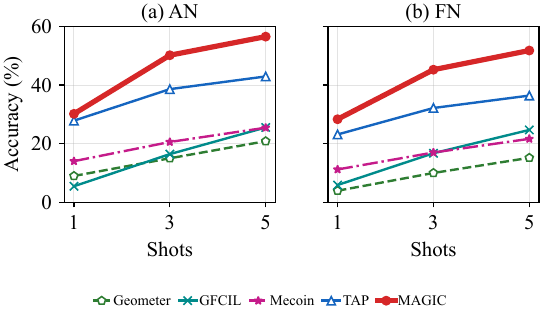}
    \caption{Novel-class performance on CoraFull.}
    \label{fig:novel_shot_performance}
\end{wrapfigure}

Figure~\ref{fig:novel_shot_performance} shows novel-class performance
under the 1-, 3-, and 5-shot settings on cora-full. AN is the average novel-class accuracy across incremental sessions, while FN measures the accuracy of
all encountered novel classes after the final session. Both metrics
generally improve as additional supports become available. MAGIC
consistently achieves the strongest performance, with its advantage
becoming more pronounced under the 3- and 5-shot settings. This demonstrates its ability to exploit limited supervision while retaining novel-class
knowledge throughout the incremental process. More data about Novel Session Performance are provided in Appendix~\ref{app:novel_class_results}.

\section{Related Work}
\label{sec:related_work}

\paragraph{Continual Graph Learning}
Graph representation learning has evolved through both trainable
architectures and fixed propagation schemes. GCN, GraphSAGE, and GAT
learn task-specific encoder parameters
~\citep{kipf2017gcn,hamilton2017inductive,velickovic2018graph}, whereas
SGC, S$^2$GC, personalized PageRank, and graph heat kernels provide
parameter-free graph propagation
~\citep{wu2019sgc,zhu2021ssgc,gasteiger2019ppnp,xu2019graphheat}.
More recently, graph foundation models such as AnyGraph and RGVT have
explored transferable representations across diverse graphs
~\citep{xia2026anygraph,lee2026rgvt}. These advances nevertheless
primarily consider static learning rather than sequential knowledge
acquisition on an evolving graph. Continual graph learning addresses
this setting by incorporating newly arriving graph data while
preserving previous knowledge. ER-GNN employs experience
replay~\citep{zhou2021ergnn}; CaT and PUMA compress historical
information through graph condensation
~\citep{liu2023cat,liu2025puma}; DeLoMe develops debiased memory
replay~\citep{niu2024delome}; TWP regularizes topology-sensitive
parameters~\citep{liu2021twp}; and DyGRAIN distills knowledge for
structurally affected nodes~\citep{kim2022dygrain}. CGLB provides
standardized evaluation protocols for this
setting~\citep{zhang2022cglb}.

\paragraph{Graph Few-Shot Class-Incremental Learning}
GFSCIL extends few-shot class-incremental learning
~\citep{tao2020fewshot,zhang2021cec,zhou2022fact,chi2022metafscil} to
evolving graphs. HAG-Meta employs pseudo-incremental meta-training and
hierarchical attention~\citep{tan2022gfcil}; Geometer learns
geometrically constrained prototypes~\citep{lu2022geometer}; Mecoin
maintains structured prototype memories that interact with a
GNN~\citep{li2024mecoin}; and TAP iteratively calibrates novel-class
prototypes using pseudo-labeled query nodes in the inductive
setting~\citep{li2025inductive}.

\paragraph{Analytic Continual Learning}
Analytic continual learning replaces gradient-based classifier updates
with closed-form solutions based on accumulated sufficient statistics.
ACIL enables replay-free analytic class-incremental
learning~\citep{zhuang2022acil}, while GKEAL extends this paradigm to
few-shot incremental learning through Gaussian kernel
embeddings~\citep{zhuang2023gkeal}. In the graph domain, GraphKeeper
employs a recursively updated ridge-regression classifier to preserve
decision boundaries across domains in graph domain-incremental
learning~\citep{guo2025graphkeeper}.

\section{Conclusion}

We presented MAGIC, a replay-free framework for graph few-shot
class-incremental learning. MAGIC combines frozen graph representation backbone,
analytic continual learning, drift-aware distillation, and
MRF-based topology-prior injection to address catastrophic forgetting,
topology-induced representation drift, and novel-class overfitting.
Extensive experiments demonstrate that MAGIC achieves strong accuracy
and training efficiency across multiple GFSCIL benchmarks.

\subsection*{AI use statement}
In this work, we used generative AI tools for providing critical ingredients for proving mathematical claims, assisting in the writing of proofs, and assisting with translation. We have not used generative AI tools for helping develop theoretical models or conceptual frameworks, formulating mathematical claims, proposing or refining hypotheses, designing or providing feedback on research methodology or experiments, implementing methods, cleaning and reformatting datasets, supporting qualitative and thematic data analysis, or interpreting results, and generating synthetic data sets is not applicable to this work. Additionally, we used generative AI tools for drafting parts of the research paper, summarizing or analysing existing literature, editing the paper to improve readability, and identifying relevant literature. We have reviewed all AI-assisted work. Specifically, all mathematical claims, proof steps, and derivations suggested or assisted by AI were independently verified line-by-line by the authors for correctness. Furthermore, all AI-identified or summarized references were manually cross-checked against the original literature to prevent hallucinated citations, and all translated/edited text was thoroughly reviewed for clarity, accuracy, and tone. We take responsibility for the final content of this work, including text, claims or artifacts produced with the aid of generative AI.

\subsection*{Reproducibility statement}
We have taken several measures to ensure the reproducibility of this
work. The GFSCIL setting is provided in Section~\ref{sec:preliminaries}, analytic update rules, relational prior,
MRF inference procedure, and drift distillation mechanism are specified
in Section~\ref{sec:methodology}. Complete proofs of the theoretical
results, together with their probabilistic interpretation, are provided
in Appendix~\ref{app:theoretical_analysis}. Dataset statistics, evaluation metrics,
and the main experimental protocol are described in
Section~\ref{sec:experiments}. Implementation details, software
environment, random seeds, and per-dataset hyperparameters are reported
in Appendix~\ref{app:reproducibility}. Additional results, including
additinal main results, and
session-wise novel-class performance, hyperparameter sensitivity analyses, are included in Appendix~\ref{sec:further_experiments}.
Code is available at \url{https://github.com/JingLiu-1234/MAGIC-for-GFSCIL}.

\bibliography{iclr2027_conference}
\bibliographystyle{iclr2027_conference}

\appendix

\section{Theoretical Analysis}
\label{app:theoretical_analysis}

This appendix provides a probabilistic interpretation of the relational
prior, Potts MRF inference, novel-class posterior projection,
conservative class balancing, and analytic reinjection used by MAGIC.
We then prove the topology-induced perturbation result and the
convergence of damped mean-field inference.

\subsection{Probabilistic Interpretation}
\label{app:probabilistic_interpretation}

\subsubsection{The Base Relation Prior as a Balanced Log Bayes Factor}
\label{app:relation_probability}

Let $S_{ij}=\mathbf 1\{y_i=y_j\}$ indicate whether two labeled base
nodes share the same class. For similarity interval $\mathcal I_b$, let
$R_{ij}^{(b)}$ denote the event that $(i,j)$ is connected and
$\mathbf h_i^\top\mathbf h_j\in\mathcal I_b$. Therefore, $R_{ij}^{(b)}$
jointly describes graph connectivity and representation similarity,
rather than similarity alone.

The raw number of same-class edges can be dominated by classes with
many labeled nodes. MAGIC removes this dependence through a balanced
pair-sampling interpretation. To sample a same-class pair, we first
select a base class uniformly and then select an unordered pair
uniformly from that class. Consequently, a pair from class $c$ has
probability $1/(|\mathcal C_0|\binom{n_c}{2})$, which is exactly the
same-class weight $w_{ij}^{+}$ used in the base-stage statistics.

Similarly, to sample a different-class pair, we first select an
unordered pair of distinct classes uniformly and then select one node
uniformly from each class. A pair from classes $c$ and $c'$ therefore
has probability
$1/(\binom{|\mathcal C_0|}{2}n_cn_{c'})$, corresponding to
$w_{ij}^{-}$.

Under these two balanced sampling distributions, define
$\pi_b^{+}=P_{\mathrm{bal}}(R_{ij}^{(b)}\mid S_{ij}=1)$ and
$\pi_b^{-}=P_{\mathrm{bal}}(R_{ij}^{(b)}\mid S_{ij}=0)$. The weighted
masses $M_b^{+}$ and $M_b^{-}$ in Section~\ref{sec:base_stage} are
empirical estimates of $\pi_b^{+}$ and $\pi_b^{-}$, respectively. The
population relation coefficient is therefore
\begin{equation}
\label{eq:population_relation}
    \beta_b^\star
    =
    \log
    \frac{
        P_{\mathrm{bal}}(R_{ij}^{(b)}\mid S_{ij}=1)
    }{
        P_{\mathrm{bal}}(R_{ij}^{(b)}\mid S_{ij}=0)
    }
    =
    \log\frac{\pi_b^{+}}{\pi_b^{-}}.
\end{equation}

By Bayes' rule, the posterior odds of label agreement satisfy
\begin{equation}
\label{eq:relation_bayes_rule}
    \log
    \frac{
        P_{\mathrm{bal}}(S_{ij}=1\mid R_{ij}^{(b)})
    }{
        P_{\mathrm{bal}}(S_{ij}=0\mid R_{ij}^{(b)})
    }
    =
    \log
    \frac{
        P_{\mathrm{bal}}(S_{ij}=1)
    }{
        P_{\mathrm{bal}}(S_{ij}=0)
    }
    +
    \beta_b^\star.
\end{equation}
Thus, $\beta_b^\star$ measures how much observing an edge in similarity
interval $\mathcal I_b$ changes the log-odds that its endpoints have
the same label. Equation~\ref{eq:relation_prior} is a smoothed and
clipped empirical estimate of this quantity. The smoothing constant
prevents an undefined ratio when one type of edge is absent, while
clipping limits the influence of unreliable estimates obtained from
sparsely populated intervals.

To derive the corresponding pairwise potential, suppose the relational
event depends on a pair of labels only through whether they agree. That
is, $P(R_{ij}^{(b)}\mid y_i=a,y_j=c)=\pi_b^{+}$ when $a=c$, and equals
$\pi_b^{-}$ otherwise. This likelihood can be rewritten as
\begin{equation}
\label{eq:potts_from_likelihood}
    P(R_{ij}^{(b)}\mid y_i=a,y_j=c)
    =
    \pi_b^{-}
    \exp
    \left(
        \beta_b^\star\mathbf 1\{a=c\}
    \right).
\end{equation}
The factor $\pi_b^{-}$ is independent of the candidate labels and is
absorbed into the normalization constant. The remaining term is
precisely a Potts pairwise potential.

When $\beta_b^\star>0$, the observed relation is more likely under label
agreement and the potential favors equal labels. When
$\beta_b^\star<0$, the relation is more likely under label disagreement
and the potential discourages equal labels. Hence, the signed prior can
express both homophilous and heterophilous relations.

For any permutation $\sigma$ of the class identities,
$\mathbf 1\{y_i=y_j\}=\mathbf 1\{\sigma(y_i)=\sigma(y_j)\}$. The prior
therefore remains unchanged under class permutation. Its transfer from
base to novel classes relies on the relation-stability assumption
$P_t(R^{(b)}\mid S)=P_0(R^{(b)}\mid S)$: the feature distributions or
semantic identities of the classes may change, but the conditional
relationship between graph relations and label agreement should remain
approximately stable.

\subsubsection{From Analytic Scores to Probabilities}
\label{app:score_probability}

The analytic classifier produces an unconstrained score vector
$\mathbf s_i=\mathbf h_i^\top\mathbf W$, which does not necessarily lie
on the probability simplex. MAGIC converts it into a valid probability
vector using Euclidean simplex projection:
\begin{equation}
\label{eq:simplex_probability}
\begin{gathered}
\Pi_{\Delta}(\mathbf s_i)
=
\arg\min_{\mathbf p\in\Delta^{C-1}}
\frac{1}{2}\|\mathbf p-\mathbf s_i\|_2^2,
\\
\Delta^{C-1}
=
\left\{
\mathbf p\in\mathbb R_{\geq0}^{C}:
\mathbf 1^\top\mathbf p=1
\right\},
\qquad
\mathbf p_{0,i}
=
\Pi_{\Delta}
\left(
\mathbf h_{t,i}^{\top}\mathbf W_t^{(0)}
\right).
\end{gathered}
\end{equation}
The projection is the unique probability vector closest to the
unconstrained analytic scores under the squared Euclidean geometry used
by the ridge-regression objective. In implementation, its entries are
lower-bounded by a small positive constant and renormalized to avoid
undefined logarithms during MRF inference.

\subsubsection{The Potts MRF Posterior}
\label{app:potts_probability}

Let $\mathbf p_{0,i}$ be the full-class probability obtained from the
naive analytic classifier through
Equation~\ref{eq:simplex_probability}. MAGIC defines the unary potential
as $\phi_i(c)=p_{0,i}(c)^{\tau_u}$, where $\tau_u$ controls the
confidence placed in the naive classifier. For a labeled support node,
the unary potential is replaced by
$\phi_i(c)=\mathbf 1\{c=y_i\}$, thereby clamping its posterior to the
observed label.

For edge $(i,j)$, let
$J_{ij}=\beta_{b(i,j)}/\sqrt{d_id_j}$. The pairwise potential is
$\psi_{ij}(a,c)=\exp(\eta J_{ij}\mathbf 1\{a=c\})$, where $\eta$
controls the global strength of the transferred prior. Combining all
unary and pairwise potentials gives
\begin{equation}
\label{eq:appendix_mrf_distribution}
    p(\mathbf y\mid\mathcal G_t,\mathbf H_t,\mathcal S_t)
    =
    \frac{1}{Z}
    \prod_i p_{0,i}(y_i)^{\tau_u}
    \prod_{(i,j)\in\mathcal E_t^{\mathrm{sub}}}
    \exp
    \left(
        \eta J_{ij}\mathbf 1\{y_i=y_j\}
    \right),
\end{equation}
which is the normalized form of
Equation~\ref{eq:mrf_distribution}. Here, $Z$ is the partition function
obtained by summing the unnormalized probability over all assignments
of the unclamped labels.

Without degree normalization and global scaling, the ratio between the
equal-label and different-label pairwise potentials would be
$\exp(\beta_b)$, corresponding directly to the estimated Bayes factor.
With normalization, this ratio becomes
$\exp(\eta\beta_b/\sqrt{d_id_j})$. Thus, MAGIC can be interpreted as
using a tempered or power likelihood: the direction of the relational
evidence is preserved, while its magnitude is adjusted to prevent
high-degree nodes from contributing the same evidence repeatedly
through many incident edges.

The unary and pairwise evidence have distinct roles. The unary term
specifies which particular class is plausible according to the naive
classifier. The pairwise term does not contain a class-specific
prototype and only evaluates whether neighboring labels should agree.
The MRF posterior therefore combines class-specific predictions with a
class-permutation-invariant relation model.

\subsubsection{Mean-Field Inference as Variational Optimization}
\label{app:mean_field_probability}

Exact marginalization of Equation~\ref{eq:appendix_mrf_distribution} is
generally intractable. MAGIC uses the factorized approximation
$q(\mathbf y)=\prod_{i\in\mathcal V_{\mathrm{free}}}q_i(y_i)$, while
the support-node distributions remain fixed one-hot vectors.

Mean-field inference minimizes
$\operatorname{KL}(q(\mathbf y)\|p(\mathbf y))$. After discarding terms
that do not depend on $q$, the variational free energy is
\begin{align}
\label{eq:variational_free_energy}
    \mathcal F(q)
    =&
    \sum_{i\in\mathcal V_{\mathrm{free}}}
    \sum_c q_i(c)\log q_i(c)
    -
    \tau_u
    \sum_{i\in\mathcal V_{\mathrm{free}}}
    \sum_c q_i(c)\log p_{0,i}(c)
    \nonumber\\
    &-
    \eta
    \sum_{(i,j)\in\mathcal E_t^{\mathrm{sub}}}
    J_{ij}
    \sum_c q_i(c)q_j(c).
\end{align}
The first term is the negative entropy of the variational distribution.
The second encourages agreement with the naive classifier, while the
third encourages or discourages neighboring label agreement according
to the sign of $J_{ij}$.

To minimize $\mathcal F$ with respect to $q_i$, introduce a Lagrange
multiplier $\lambda_i$ for the constraint $\sum_cq_i(c)=1$. The partial
derivative is
$\partial\mathcal F/\partial q_i(c)=\log q_i(c)+1-\tau_u\log
p_{0,i}(c)-\eta\sum_{j\in\mathcal N_i}J_{ij}q_j(c)+\lambda_i$.
Setting it to zero gives
$\log q_i(c)=\tau_u\log p_{0,i}(c)+
\eta\sum_{j\in\mathcal N_i}J_{ij}q_j(c)-1-\lambda_i$.

Exponentiating this equality and normalizing over the classes yields
\begin{equation}
\label{eq:mean_field_derivation}
    q_i(c)
    =
    \frac{
        p_{0,i}(c)^{\tau_u}
        \exp\left(
            \eta\sum_{j\in\mathcal N_i}J_{ij}q_j(c)
        \right)
    }{
        \sum_{c'}
        p_{0,i}(c')^{\tau_u}
        \exp\left(
            \eta\sum_{j\in\mathcal N_i}J_{ij}q_j(c')
        \right)
    }.
\end{equation}
This is equivalent to the softmax proposal in
Equation~\ref{eq:mean_field}. Damping replaces the proposal
$\widehat{\mathbf q}_i$ by the convex combination
$(1-\alpha)\mathbf q_i+\alpha\widehat{\mathbf q}_i$. Since
$\alpha>0$, a distribution is a fixed point of the damped update if and
only if it is a fixed point of the original mean-field update.

\subsubsection{An EM-Like Interpretation of Novel-Stage Learning}
\label{app:em_interpretation}
The two-step novel-stage update follows an EM-like
predict--refine--update procedure~\citep{dempster1977maximum}. Given the
few-shot supports and distillation targets, MAGIC first obtains the
naive classifier $\mathbf W_t^{0}$. Analogous to an E-step, it then
treats the labels of unlabeled candidate nodes as latent variables and
uses Mean-Field inference to estimate their graph-refined posterior
distributions $\mathbf q_i$. Analogous to an M-step, these posteriors
are converted into supervision
$\boldsymbol{\alpha}_i$ and inserted into the analytic sufficient
statistics, after which $\mathbf W_t^{1}$ is obtained in closed form.
The overall update can therefore be summarized as
$\mathbf W_t^{0}\rightarrow\{\mathbf q_i\}\rightarrow
\{\boldsymbol{\alpha}_i\}\rightarrow\mathbf W_t^{1}$.

\subsubsection{Unified Probabilistic Interpretation of Analytic Injection}
\label{app:analytic_probability}

Let $\boldsymbol{\gamma}_i\in\mathbb R_{\geq0}^{C}$ be any nonnegative
supervision vector. It may represent a balanced hard label, a
distillation target, or an MRF pseudo-label. Define its total mass as
$m_i=\|\boldsymbol{\gamma}_i\|_1$. If $m_i>0$, its normalized target is
$\mathbf p_i=\boldsymbol{\gamma}_i/m_i$.

Consider the Gaussian working likelihood
\begin{equation}
\label{eq:analytic_likelihood}
    \mathbf p_i
    \mid\mathbf h_i,\mathbf W,m_i
    \sim
    \mathcal N
    \left(
        \mathbf h_i^\top\mathbf W,
        \frac{1}{2m_i}\mathbf I
    \right).
\end{equation}
A larger supervision mass corresponds to lower observation variance and
therefore greater confidence. Assume the Gaussian parameter prior
$p(\mathbf W)\propto\exp(-\lambda\|\mathbf W\|_F^2)$. Ignoring constants
independent of $\mathbf W$, the negative log-posterior becomes
\begin{equation}
\label{eq:analytic_map_objective}
    \mathcal L(\mathbf W)
    =
    \sum_i
    m_i
    \left\|
        \mathbf h_i^\top\mathbf W-\mathbf p_i
    \right\|_2^2
    +
    \lambda\|\mathbf W\|_F^2.
\end{equation}

Since $m_i\mathbf p_i=\boldsymbol{\gamma}_i$, differentiating with
respect to $\mathbf W$ and setting the derivative to zero gives
\begin{equation}
\label{eq:analytic_normal_equation}
    \left(
        \sum_i m_i\mathbf h_i\mathbf h_i^\top
        +
        \lambda\mathbf I
    \right)\mathbf W
    =
    \sum_i
    \mathbf h_i\boldsymbol{\gamma}_i^\top.
\end{equation}
Thus, defining
$\mathbf R=\sum_i m_i\mathbf h_i\mathbf h_i^\top$ and
$\mathbf C=\sum_i\mathbf h_i\boldsymbol{\gamma}_i^\top$ recovers
$\mathbf W=(\mathbf R+\lambda\mathbf I)^{-1}\mathbf C$, consistent with
Equation~\ref{eq:base_analytic}.

This result explains why the three supervision sources are
mathematically compatible. A true one-hot label is a probability target
with explicit class-balanced precision. A distillation distribution is
a soft observation that preserves the historical decision function.
An MRF posterior is a graph-refined soft observation whose supervision
mass is determined by novel-class posterior projection and conservative
class balancing. The retained novel-class mass controls each node's
contribution, while class-level scaling limits the aggregate
pseudo-supervision mass to at most $\rho$ per novel class.

\subsection{Proofs of Theoretical Results}
\label{app:theorem_proofs}

\subsubsection{Proof of Theorem~\ref{thm:topology_drift}}
\label{app:proof_topology_drift}

Let $\mathbf H^{+}=\mathbf H^{-}+\boldsymbol{\Delta}$. Before graph
evolution, the historical nodes contribute
$\mathbf R^{-}=(\mathbf H^{-})^\top\mathbf D\mathbf H^{-}$ and
$\mathbf C^{-}=(\mathbf H^{-})^\top\boldsymbol{\Gamma}$. After graph
evolution,
\begin{align}
    \mathbf R^{+}
    &=
    (\mathbf H^{-}+\boldsymbol{\Delta})^\top
    \mathbf D
    (\mathbf H^{-}+\boldsymbol{\Delta})
    \nonumber\\
    &=
    \mathbf R^{-}
    +
    (\mathbf H^{-})^\top\mathbf D\boldsymbol{\Delta}
    +
    \boldsymbol{\Delta}^\top\mathbf D\mathbf H^{-}
    +
    \boldsymbol{\Delta}^\top\mathbf D\boldsymbol{\Delta},
\end{align}
and
$\mathbf C^{+}
=(\mathbf H^{-}+\boldsymbol{\Delta})^\top\boldsymbol{\Gamma}
=\mathbf C^{-}+\boldsymbol{\Delta}^\top\boldsymbol{\Gamma}$.
Subtracting the previous statistics proves both identities.
\hfill$\square$

\subsubsection{Proof of Theorem~\ref{thm:mean_field_convergence}}
\label{app:proof_mean_field}

Only free nodes need to be considered because support-node
distributions are clamped. Messages from support nodes are constant and
can therefore be absorbed into a unary matrix $\mathbf U$. Let
$\mathbf Q\in\mathbb R^{n_f\times C}$ contain the free-node
distributions and let $\mathbf J_{\mathrm{FF}}$ be their coupling
matrix. The undamped mean-field mapping is
\begin{equation}
\label{eq:undamped_mapping}
    \mathcal T(\mathbf Q)
    =
    \operatorname{Softmax}_{\mathrm{row}}
    \left(
        \mathbf U+\eta\mathbf J_{\mathrm{FF}}\mathbf Q
    \right).
\end{equation}

We first derive a Lipschitz bound for softmax. For
$\mathbf p=\operatorname{softmax}(\mathbf z)$, its Jacobian is
$\mathbf J_{\mathrm{sm}}(\mathbf p)=
\operatorname{diag}(\mathbf p)-\mathbf p\mathbf p^\top$. This matrix is
symmetric and positive semidefinite. In row $c$, the diagonal entry is
$p_c(1-p_c)$ and the sum of the absolute off-diagonal entries is also
$p_c(1-p_c)$.

By the Gershgorin circle theorem, every eigenvalue of
$\mathbf J_{\mathrm{sm}}$ is no larger than
$2p_c(1-p_c)$ for some $c$. Since
$p_c(1-p_c)\leq1/4$, it follows that
$\|\mathbf J_{\mathrm{sm}}(\mathbf p)\|_2\leq1/2$. The mean-value theorem
therefore gives
\begin{equation}
\label{eq:softmax_lipschitz}
    \|
        \operatorname{softmax}(\mathbf z)
        -
        \operatorname{softmax}(\mathbf z')
    \|_2
    \leq
    \frac{1}{2}\|\mathbf z-\mathbf z'\|_2.
\end{equation}

Applying Equation~\ref{eq:softmax_lipschitz} independently to every row
of Equation~\ref{eq:undamped_mapping} yields
\begin{align}
    \|\mathcal T(\mathbf Q)-\mathcal T(\mathbf Q')\|_F
    &\leq
    \frac{\eta}{2}
    \|
        \mathbf J_{\mathrm{FF}}
        (\mathbf Q-\mathbf Q')
    \|_F
    \nonumber\\
    &\leq
    \frac{\eta}{2}
    \|\mathbf J_{\mathrm{FF}}\|_2
    \|\mathbf Q-\mathbf Q'\|_F.
\end{align}

The damped mapping is
$\mathcal T_\alpha(\mathbf Q)=
(1-\alpha)\mathbf Q+\alpha\mathcal T(\mathbf Q)$. Thus,
\begin{align}
    \|
        \mathcal T_\alpha(\mathbf Q)
        -
        \mathcal T_\alpha(\mathbf Q')
    \|_F
    \leq&
    (1-\alpha)\|\mathbf Q-\mathbf Q'\|_F
    \nonumber\\
    &+
    \alpha
    \|\mathcal T(\mathbf Q)-\mathcal T(\mathbf Q')\|_F
    \nonumber\\
    \leq&
    \left(
        1-\alpha
        +
        \frac{\alpha\eta}{2}
        \|\mathbf J_{\mathrm{FF}}\|_2
    \right)
    \|\mathbf Q-\mathbf Q'\|_F.
\end{align}
Define
$\kappa_{\mathrm{MF}}=
1-\alpha+\alpha\eta\|\mathbf J_{\mathrm{FF}}\|_2/2$. Under the condition
$\eta\|\mathbf J_{\mathrm{FF}}\|_2/2<1$, we have
$\kappa_{\mathrm{MF}}<1$, so the damped mapping is a contraction.

The product of node-wise probability simplices is a closed subset of a
finite-dimensional Euclidean space and is therefore complete. By the
Banach fixed-point theorem, the mapping has a unique fixed point
$\mathbf Q^\star$. Repeatedly applying the contraction inequality gives
\begin{equation}
\label{eq:mean_field_geometric_convergence}
    \|\mathbf Q^{(\ell)}-\mathbf Q^\star\|_F
    \leq
    \kappa_{\mathrm{MF}}^\ell
    \|\mathbf Q^{(0)}-\mathbf Q^\star\|_F.
\end{equation}
Therefore, the damped mean-field iterates converge geometrically to the
unique fixed point, completing the proof of
Theorem~\ref{thm:mean_field_convergence}.
\hfill$\square$

\section{Further Experiments}
\label{sec:further_experiments}

\subsection{Datasets}
\label{sec:datasets}
We evaluate MAGIC on five widely used node-classification benchmarks:
CoraFull~\citep{bojchevski2018deep}, Coauthor
CS~\citep{shchur2018pitfalls}, ogbn-arxiv~\citep{hu2020ogb},
WikiCS~\citep{mernyei2020wikics}, and Amazon
Computers~\citep{shchur2018pitfalls}. CoraFull and ogbn-arxiv are
citation networks in which nodes represent papers and labels correspond
to research topics. Coauthor CS connects authors who have coauthored
papers, with labels indicating their primary research fields. WikiCS is
a hyperlink network of computer-science articles from Wikipedia, while
Amazon Computers is a product co-purchase network whose labels denote
product categories.
Table~\ref{tab:dataset_statistics} summarizes the dataset statistics.
We report edge homophily
$h_{\mathrm{edge}}=|\{(u,v)\in\mathcal E:y_u=y_v\}|/|\mathcal E|$,
which measures the proportion of edges connecting nodes with the same
label.
\begin{table}[t]
    \caption{Statistics of the node-classification datasets.}
    \label{tab:dataset_statistics}
    \centering
    \scriptsize
    \setlength{\tabcolsep}{2.5pt}
    \renewcommand{\arraystretch}{1.4}
    \begin{tabular}{L{2.80cm}*{5}{C{1.00cm}}}
        \toprule
        Dataset & Nodes & Edges & Features & Classes &
        $h_{\mathrm{edge}}$ \\
        \midrule
        CoraFull         & 19,793  & 126,842   & 8,710 & 70 & 0.57 \\
        Coauthor CS      & 18,333  & 163,788   & 6,805 & 15 & 0.81 \\
        ogbn-arxiv       & 169,343 & 1,166,243 & 128   & 40 & 0.65 \\
        WikiCS           & 11,701  & 216,123   & 300   & 10 & 0.65 \\
        Amazon Computers & 13,752  & 491,722   & 767   & 10 & 0.78 \\
        \bottomrule
    \end{tabular}
\end{table}

\subsection{Class-Incremental Splits}
\label{app:class-incremental_splits}

Table~\ref{tab:class_splits} reports the number of base classes and
novel classes introduced per incremental session for each dataset.

\begin{table}[t]
    \caption{Class-incremental split configurations.}
    \label{tab:class_splits}
    \centering
    \scriptsize
    \setlength{\tabcolsep}{10pt}
    \renewcommand{\arraystretch}{1.15}
    \begin{tabular}{L{2.80cm}C{1.60cm}C{1.80cm}}
        \toprule
        Dataset & Base Classes & Way \\
        \midrule
        CoraFull         & 20 & 5 \\
        ogbn-arxiv       & 10 & 3 \\
        Coauthor CS      & 5  & 2 \\
        Amazon Computers & 4  & 2 \\
        WikiCS           & 4  & 2 \\
        \bottomrule
    \end{tabular}
\end{table}

\subsection{Compared Baselines}
\label{app:baselines}

We compare MAGIC with four representative continual-learning methods
and four GFSCIL-specific methods. All baselines are evaluated using the
same class splits, support sets, visible graphs, and evaluation protocol
as MAGIC.

\paragraph{LwF.}
Learning without Forgetting (LwF) preserves previous knowledge through
knowledge distillation~\citep{li2016learning}. Before learning a new
session, it records the predictions of the previous model and uses them
as soft targets while optimizing the current model. It does not require
storing historical samples, but its effectiveness depends on whether
the newly available data can provide sufficient coverage of previously
learned classes.

\paragraph{EWC.}
Elastic Weight Consolidation (EWC) estimates the importance of each
model parameter using the diagonal of the Fisher information
matrix~\citep{kirkpatrick2017overcoming}. During subsequent sessions, it
adds a quadratic penalty that discourages important parameters from
moving away from their previous values. EWC primarily protects
historical knowledge at the parameter level and does not explicitly
address the scarcity of novel-class supervision.

\paragraph{MAS.}
Memory Aware Synapses (MAS) measures parameter importance according to
the sensitivity of the model output to changes in each
parameter~\citep{aljundi2018memory}. Unlike EWC, its importance
estimation does not require labels and can therefore exploit unlabeled
inputs. Important parameters are subsequently protected by a weighted
quadratic regularizer during incremental learning.

\paragraph{TWP.}
Topology-aware Weight Preserving (TWP) is a continual graph-learning
method specifically designed for GNNs~\citep{liu2021twp}. In addition
to measuring parameter importance for node classification, TWP
identifies parameters important to neighborhood aggregation and
topological information. It regularizes these parameters in later
sessions to mitigate forgetting of both predictive and structural
knowledge.

\paragraph{GFCIL.}
The original GFSCIL framework introduces graph pseudo-incremental
learning and the HAG-Meta model~\citep{tan2022gfcil}. It constructs
pseudo-incremental episodes from base classes to simulate future
few-shot sessions and employs hierarchical attention at the task and
node levels. A task-sensitive regularizer is further used to balance
adaptation to novel classes and retention of base knowledge.

\paragraph{Geometer.}
Geometer represents each class through an attention-based prototype and
imposes geometric constraints on the evolving prototype
space~\citep{lu2022geometer}. Its prototype representation incorporates
graph information to improve class separability, while knowledge
distillation helps preserve previously learned classes. The resulting
classifier performs incremental prediction by comparing node
representations with the accumulated class prototypes.

\paragraph{Mecoin.}
Mecoin introduces a structured memory unit for efficiently storing and
updating class prototypes~\citep{li2024mecoin}. Its memory-construction
module enables interactions between incoming nodes and stored
prototypes, while a memory-representation adaptation module separates
prototype learning from class prediction. It further exchanges
knowledge between the memory and GNN to reduce extensive encoder
fine-tuning and preserve historical information.

\paragraph{TAP.}
TAP addresses inductive GFSCIL through topology-based class augmentation
and prototype calibration~\citep{li2025inductive}. It augments the
limited novel-class supports using graph-topological information and
iteratively calibrates novel prototypes with confident pseudo-labeled
query nodes. This improves few-shot generalization while retaining a
prototype-based incremental classifier.

\subsection{Backbone Implementations}
\label{app:backbone_details}

We evaluate five frozen graph representation backbones. For the four
parameter-free operators, let
$\mathbf P_t=\widetilde{\mathbf D}_t^{-1/2}
\widetilde{\mathbf A}_t\widetilde{\mathbf D}_t^{-1/2}$ denote the
symmetrically normalized adjacency matrix with self-loops, and let
$\mathbf X'_t$ denote the node attributes after the optional fixed
random projection. All resulting node representations are
$\ell_2$-normalized row-wise before being used by MAGIC.

\paragraph{SGC.}
Simplified Graph Convolution (SGC) removes intermediate nonlinearities
and trainable transformations, retaining only fixed graph propagation
followed by a linear classifier~\citep{wu2019sgc}. Its representation is
\begin{equation}
    \overline{\mathbf H}_t
    =
    \mathbf P_t^K\mathbf X'_t ,
\end{equation}
where $K$ is both the propagation depth and the maximum receptive-field
radius. SGC retains only the representation obtained after the final
propagation step.

\paragraph{S\textsuperscript{2}GC.}
Simple Spectral Graph Convolution (S\textsuperscript{2}GC) combines
representations from multiple propagation depths rather than retaining
only the $K$th-order result~\citep{zhu2021ssgc}. We implement
\begin{equation}
    \overline{\mathbf H}_t
    =
    \alpha\mathbf X'_t
    +
    \frac{1-\alpha}{K}
    \sum_{\ell=1}^{K}
    \mathbf P_t^\ell\mathbf X'_t ,
\end{equation}
where $\alpha\in[0,1]$ controls the residual contribution of the
original attributes. This operator averages, rather than concatenates,
the different propagation orders and has a maximum range of $K$ hops.

\paragraph{PPR.}
Personalized PageRank (PPR) propagation repeatedly diffuses node
features while restarting from the original attributes
~\citep{gasteiger2019ppnp}. Starting from
$\mathbf H_t^{(0)}=\mathbf X'_t$, our finite-step implementation uses
\begin{equation}
    \mathbf H_t^{(\ell+1)}
    =
    (1-\alpha)\mathbf P_t\mathbf H_t^{(\ell)}
    +
    \alpha\mathbf X'_t,
    \qquad \ell=0,\ldots,K-1,
\end{equation}
and returns $\overline{\mathbf H}_t=\mathbf H_t^{(K)}$. Here, $\alpha$
is the restart probability. Unlike the exact infinite-step PPR
solution, this truncated recurrence requires only $K$ sparse
propagations and restricts the influence range to at most $K$ hops.

\paragraph{Truncated Heat Kernel.}
Heat-kernel diffusion propagates graph signals continuously through
$\exp[-\tau(\mathbf I-\mathbf P_t)]$~\citep{xu2019graphheat}. We avoid
matrix exponentiation by truncating its Taylor expansion:
\begin{equation}
    \overline{\mathbf H}_t
    =
    \sum_{\ell=0}^{K}
    w_\ell\mathbf P_t^\ell\mathbf X'_t,
    \qquad
    w_\ell
    =
    \frac{\tau^\ell/\ell!}
    {\sum_{r=0}^{K}\tau^r/r!}.
\end{equation}
The diffusion time $\tau\geq0$ determines how strongly higher-order
neighborhoods contribute, while $K$ controls the truncation order and
strict maximum hop range. The common factor $\exp(-\tau)$ cancels under
coefficient and row normalization.

\paragraph{RGVT.}
Recurrent Graph View Transformation (RGVT) is a pretrained graph
representation model that applies a shared scalar transformation to
multi-hop graph views~\citep{lee2026rgvt}. Let
$\mathbf P_t^{\mathrm{rw}}$ and $\mathbf P_t^{\mathrm{sym}}$ denote the
random-walk and symmetric propagation operators, respectively. For
node $i$ and feature channel $f$, the released encoder constructs
\begin{equation}
    \mathbf v_{i,f}(\mathbf H)
    =
    \left[
    H_{i,f},
    (\mathbf P_t^{\mathrm{rw}}\mathbf H)_{i,f},
    (\mathbf P_t^{\mathrm{sym}}\mathbf H)_{i,f},
    ((\mathbf P_t^{\mathrm{rw}})^2\mathbf H)_{i,f},
    ((\mathbf P_t^{\mathrm{sym}})^2\mathbf H)_{i,f}
    \right]
\end{equation}
and computes
$H_{i,f}^{(r+1)}=g_{\boldsymbol{\theta}}
(\mathbf v_{i,f}(\mathbf H^{(r)}))$ using the same pretrained MLP
$g_{\boldsymbol{\theta}}$ for every node and feature channel. MAGIC
freezes $\boldsymbol{\theta}$ and applies this transformation
recurrently for $R$ rounds, producing an effective receptive field of
at most $2R$ hops. The released checkpoint fixes the view order and MLP
depth to two. We exclude RGVT from ogbn-arxiv because the checkpoint was
supervised-pretrained on that dataset, which would cause information
leakage.

\subsection{Additional Main Results}
\label{app:additional_main_results}

Tables~\ref{tab:additional_1shot_all} and
\ref{tab:additional_3shot_all} report the complete 1- and 3-shot
results on all five datasets. The corresponding 5-shot results are
presented in the main text.

\begin{table}[t]
    \caption{Main results (\%) under the 1-shot setting. The mean and
    standard deviation over five runs are reported in consecutive rows.
    The best result in each column is shown in bold. MAGIC Backbone: SGC.}
    \label{tab:additional_1shot_all}
    \centering
    \scriptsize
    \setlength{\tabcolsep}{1.2pt}
    \renewcommand{\arraystretch}{1.08}
    \begin{tabular}{L{1.70cm}*{15}{C{0.70cm}}}
        \toprule
        & \multicolumn{3}{c}{CoraFull}
        & \multicolumn{3}{c}{ogbn-arxiv}
        & \multicolumn{3}{c}{Coauthor CS}
        & \multicolumn{3}{c}{Amazon Computers}
        & \multicolumn{3}{c}{WikiCS} \\
        \cmidrule(lr){2-4}
        \cmidrule(lr){5-7}
        \cmidrule(lr){8-10}
        \cmidrule(lr){11-13}
        \cmidrule(lr){14-16}
        Method
        & MA & FA & PD & MA & FA & PD & MA & FA & PD
        & MA & FA & PD & MA & FA & PD \\
        \midrule

        \rowcolor{meanrow}
        LwF
        & 13.18 & 2.28 & 81.39
        & 16.64 & 5.29 & 81.38
        & 35.23 & 16.68 & 80.21
        & 32.84 & 13.17 & 82.40
        & 34.56 & 11.45 & 81.14 \\
        LwF (std)
        & $\pm$0.93 & $\pm$0.93 & $\pm$2.91
        & $\pm$1.53 & $\pm$6.77 & $\pm$7.35
        & $\pm$3.34 & $\pm$5.40 & $\pm$5.82
        & $\pm$2.10 & $\pm$5.34 & $\pm$4.25
        & $\pm$2.40 & $\pm$4.66 & $\pm$4.76 \\

        \rowcolor{meanrow}
        EWC
        & 13.53 & 2.83 & 80.85
        & 17.48 & 9.11 & 77.57
        & 36.50 & 17.14 & 79.76
        & 33.16 & 14.63 & 80.92
        & 34.83 & 11.76 & 80.91 \\
        EWC (std)
        & $\pm$1.09 & $\pm$1.71 & $\pm$2.75
        & $\pm$1.25 & $\pm$5.30 & $\pm$5.67
        & $\pm$4.02 & $\pm$5.25 & $\pm$5.64
        & $\pm$2.62 & $\pm$6.77 & $\pm$5.92
        & $\pm$2.48 & $\pm$4.98 & $\pm$5.12 \\

        \rowcolor{meanrow}
        MAS
        & 16.40 & 6.16 & 77.52
        & 30.77 & 18.51 & 68.16
        & 51.31 & 24.55 & 72.34
        & 34.88 & 12.71 & 82.95
        & 35.65 & 12.35 & 80.26 \\
        MAS (std)
        & $\pm$1.80 & $\pm$2.48 & $\pm$3.02
        & $\pm$7.41 & $\pm$3.40 & $\pm$6.17
        & $\pm$10.34 & $\pm$9.14 & $\pm$8.87
        & $\pm$1.64 & $\pm$5.01 & $\pm$4.10
        & $\pm$2.47 & $\pm$3.94 & $\pm$4.33 \\

        \rowcolor{meanrow}
        TWP
        & 15.60 & 3.98 & 79.20
        & 34.73 & 18.94 & 66.68
        & 39.70 & 16.54 & 80.34
        & 32.85 & 12.53 & 83.28
        & 34.60 & 11.41 & 80.75 \\
        TWP (std)
        & $\pm$1.06 & $\pm$1.31 & $\pm$3.12
        & $\pm$9.08 & $\pm$4.69 & $\pm$3.58
        & $\pm$3.42 & $\pm$8.27 & $\pm$8.77
        & $\pm$1.13 & $\pm$5.32 & $\pm$4.23
        & $\pm$1.94 & $\pm$4.92 & $\pm$4.75 \\

        \rowcolor{meanrow}
        GFCIL
        & 44.30 & 28.22 & 52.91
        & 47.80 & 30.00 & 52.74
        & 54.15 & 31.32 & 64.92
        & 79.05 & 64.06 & 28.58
        & 68.82 & 53.56 & 37.10 \\
        GFCIL (std)
        & $\pm$2.44 & $\pm$3.67 & $\pm$6.65
        & $\pm$8.56 & $\pm$4.75 & $\pm$3.59
        & $\pm$7.73 & $\pm$6.10 & $\pm$7.08
        & $\pm$3.42 & $\pm$6.17 & $\pm$8.39
        & $\pm$4.82 & $\pm$6.47 & $\pm$6.70 \\

        \rowcolor{meanrow}
        Geometer
        & 44.10 & 26.99 & 53.27
        & 36.31 & 19.71 & 50.50
        & 64.60 & 42.82 & 53.79
        & 73.09 & 58.54 & 32.13
        & 62.49 & 41.91 & 47.06 \\
        Geometer (std)
        & $\pm$2.23 & $\pm$3.06 & $\pm$4.38
        & $\pm$5.59 & $\pm$5.08 & $\pm$9.03
        & $\pm$8.61 & $\pm$13.23 & $\pm$13.01
        & $\pm$4.10 & $\pm$5.74 & $\pm$3.51
        & $\pm$4.04 & $\pm$7.48 & $\pm$8.97 \\

        \rowcolor{meanrow}
        Mecoin
        & 31.25 & 18.62 & 41.40
        & 24.19 & 15.63 & \textbf{31.48}
        & 57.86 & 37.38 & 54.51
        & 56.23 & 39.48 & 34.70
        & 49.33 & 35.69 & 32.40 \\
        Mecoin (std)
        & $\pm$3.83 & $\pm$1.42 & $\pm$8.39
        & $\pm$4.29 & $\pm$0.62 & \textbf{$\pm$6.77}
        & $\pm$9.39 & $\pm$7.60 & $\pm$5.46
        & $\pm$8.20 & $\pm$7.36 & $\pm$8.82
        & $\pm$9.04 & $\pm$8.03 & $\pm$5.47 \\

        \rowcolor{meanrow}
        TAP
        & 52.76 & 36.47 & 46.81
        & 54.47 & 38.62 & 46.43
        & 76.85 & 64.02 & 32.54
        & 81.02 & 68.44 & 26.79
        & 73.67 & 58.91 & 32.87 \\
        TAP (std)
        & $\pm$2.71 & $\pm$2.95 & $\pm$4.99
        & $\pm$5.82 & $\pm$2.91 & $\pm$2.27
        & $\pm$3.37 & $\pm$6.66 & $\pm$6.39
        & $\pm$6.33 & $\pm$7.82 & $\pm$7.16
        & $\pm$3.33 & $\pm$4.76 & $\pm$5.76 \\

        \rowcolor{meanrow}
        \textbf{MAGIC}
        & \textbf{54.64} & \textbf{41.99} & \textbf{38.21}
        & \textbf{55.48} & \textbf{40.71} & 43.47
        & \textbf{89.97} & \textbf{86.48} & \textbf{10.51}
        & \textbf{84.06} & \textbf{76.68} & \textbf{14.06}
        & \textbf{76.82} & \textbf{65.68} & \textbf{23.05} \\
        \textbf{MAGIC (std)}
        & \textbf{$\pm$2.17} & \textbf{$\pm$2.91}
        & \textbf{$\pm$5.36}
        & \textbf{$\pm$7.96} & \textbf{$\pm$5.37} & $\pm$3.90
        & \textbf{$\pm$3.27} & \textbf{$\pm$3.94}
        & \textbf{$\pm$4.84}
        & \textbf{$\pm$3.37} & \textbf{$\pm$5.67}
        & \textbf{$\pm$7.78}
        & \textbf{$\pm$2.88} & \textbf{$\pm$3.93}
        & \textbf{$\pm$3.95} \\
        \bottomrule
    \end{tabular}
\end{table}

\begin{table}[t]
    \caption{Main results (\%) under the 3-shot setting. The mean and
    standard deviation over five runs are reported in consecutive rows.
    The best result in each column is shown in bold. MAGIC Backbone: SGC.}
    \label{tab:additional_3shot_all}
    \centering
    \scriptsize
    \setlength{\tabcolsep}{1.2pt}
    \renewcommand{\arraystretch}{1.08}
    \begin{tabular}{L{1.70cm}*{15}{C{0.70cm}}}
        \toprule
        & \multicolumn{3}{c}{CoraFull}
        & \multicolumn{3}{c}{ogbn-arxiv}
        & \multicolumn{3}{c}{Coauthor CS}
        & \multicolumn{3}{c}{Amazon Computers}
        & \multicolumn{3}{c}{WikiCS} \\
        \cmidrule(lr){2-4}
        \cmidrule(lr){5-7}
        \cmidrule(lr){8-10}
        \cmidrule(lr){11-13}
        \cmidrule(lr){14-16}
        Method
        & MA & FA & PD & MA & FA & PD & MA & FA & PD
        & MA & FA & PD & MA & FA & PD \\
        \midrule

        \rowcolor{meanrow}
        LwF
        & 15.46 & 3.51 & 80.17
        & 18.14 & 5.29 & 81.33
        & 35.72 & 16.99 & 79.90
        & 34.54 & 13.39 & 82.23
        & 35.02 & 12.51 & 80.11 \\
        LwF (std)
        & $\pm$1.24 & $\pm$1.15 & $\pm$2.46
        & $\pm$0.66 & $\pm$6.95 & $\pm$7.53
        & $\pm$3.54 & $\pm$5.19 & $\pm$5.42
        & $\pm$2.29 & $\pm$5.47 & $\pm$4.37
        & $\pm$0.70 & $\pm$3.58 & $\pm$3.92 \\

        \rowcolor{meanrow}
        EWC
        & 15.56 & 3.61 & 80.07
        & 16.05 & 6.13 & 80.49
        & 36.66 & 17.07 & 79.82
        & 35.00 & 14.46 & 81.12
        & 35.16 & 12.57 & 80.11 \\
        EWC (std)
        & $\pm$1.36 & $\pm$1.45 & $\pm$2.43
        & $\pm$2.05 & $\pm$6.67 & $\pm$7.13
        & $\pm$4.20 & $\pm$5.07 & $\pm$5.32
        & $\pm$2.59 & $\pm$6.46 & $\pm$5.58
        & $\pm$0.80 & $\pm$3.74 & $\pm$4.14 \\

        \rowcolor{meanrow}
        MAS
        & 16.04 & 6.08 & 77.60
        & 32.89 & 27.88 & 58.80
        & 49.00 & 21.31 & 75.58
        & 34.83 & 12.54 & 83.04
        & 35.23 & 12.35 & 80.33 \\
        MAS (std)
        & $\pm$1.32 & $\pm$2.00 & $\pm$1.45
        & $\pm$5.76 & $\pm$7.97 & $\pm$7.27
        & $\pm$7.16 & $\pm$5.98 & $\pm$6.04
        & $\pm$1.61 & $\pm$5.26 & $\pm$4.14
        & $\pm$1.96 & $\pm$3.78 & $\pm$4.18 \\

        \rowcolor{meanrow}
        TWP
        & 17.01 & 5.65 & 77.47
        & 34.35 & 18.63 & 66.99
        & 40.16 & 16.45 & 80.43
        & 34.47 & 13.29 & 82.52
        & 34.38 & 12.43 & 79.74 \\
        TWP (std)
        & $\pm$2.02 & $\pm$1.84 & $\pm$2.65
        & $\pm$9.25 & $\pm$6.58 & $\pm$6.22
        & $\pm$4.58 & $\pm$3.35 & $\pm$3.25
        & $\pm$1.62 & $\pm$5.56 & $\pm$4.14
        & $\pm$0.98 & $\pm$3.76 & $\pm$3.89 \\

        \rowcolor{meanrow}
        GFCIL
        & 50.11 & 36.17 & 44.96
        & 54.56 & 35.59 & 47.19
        & 69.42 & 59.58 & 36.66
        & 85.15 & 76.36 & 16.29
        & 72.15 & 60.15 & 30.81 \\
        GFCIL (std)
        & $\pm$2.03 & $\pm$2.73 & $\pm$5.70
        & $\pm$5.27 & $\pm$5.52 & $\pm$4.82
        & $\pm$7.59 & $\pm$6.63 & $\pm$7.38
        & $\pm$3.33 & $\pm$3.07 & $\pm$5.01
        & $\pm$4.31 & $\pm$5.30 & $\pm$5.62 \\

        \rowcolor{meanrow}
        Geometer
        & 47.55 & 30.15 & 50.28
        & 39.19 & 22.45 & 47.78
        & 77.45 & 57.80 & 38.82
        & 77.68 & 62.16 & 28.61
        & 65.79 & 44.97 & 44.03 \\
        Geometer (std)
        & $\pm$1.86 & $\pm$3.08 & $\pm$4.66
        & $\pm$3.59 & $\pm$5.67 & $\pm$5.59
        & $\pm$4.64 & $\pm$6.66 & $\pm$7.13
        & $\pm$3.36 & $\pm$5.03 & $\pm$5.71
        & $\pm$4.64 & $\pm$4.31 & $\pm$4.16 \\

        \rowcolor{meanrow}
        Mecoin
        & 42.52 & 27.58 & 45.35
        & 36.86 & 23.16 & 47.17
        & 68.45 & 50.42 & 45.34
        & 67.53 & 51.24 & 36.27
        & 57.77 & 41.90 & 41.89 \\
        Mecoin (std)
        & $\pm$2.70 & $\pm$1.47 & $\pm$5.33
        & $\pm$2.07 & $\pm$0.68 & $\pm$3.82
        & $\pm$6.81 & $\pm$4.79 & $\pm$4.22
        & $\pm$1.66 & $\pm$4.30 & $\pm$4.11
        & $\pm$6.77 & $\pm$8.31 & $\pm$11.06 \\

        \rowcolor{meanrow}
        TAP
        & 57.84 & 42.88 & 40.54
        & 57.26 & 41.46 & 43.60
        & 78.52 & 66.62 & 29.83
        & 83.04 & 72.60 & 22.31
        & 75.77 & 62.69 & 29.18 \\
        TAP (std)
        & $\pm$2.29 & $\pm$1.53 & $\pm$3.17
        & $\pm$5.39 & $\pm$1.69 & $\pm$3.69
        & $\pm$3.39 & $\pm$3.94 & $\pm$3.51
        & $\pm$5.00 & $\pm$4.58 & $\pm$3.66
        & $\pm$3.35 & $\pm$5.49 & $\pm$6.54 \\

        \rowcolor{meanrow}
        \textbf{MAGIC}
        & \textbf{62.92} & \textbf{52.29} & \textbf{27.91}
        & \textbf{62.07} & \textbf{47.82} & \textbf{36.36}
        & \textbf{93.57} & \textbf{90.58} & \textbf{6.40}
        & \textbf{85.44} & \textbf{79.09} & \textbf{11.65}
        & \textbf{78.44} & \textbf{71.74} & \textbf{17.00} \\
        \textbf{MAGIC (std)}
        & \textbf{$\pm$0.79} & \textbf{$\pm$0.95}
        & \textbf{$\pm$3.81}
        & \textbf{$\pm$4.65} & \textbf{$\pm$3.00}
        & \textbf{$\pm$3.64}
        & \textbf{$\pm$0.71} & \textbf{$\pm$0.71}
        & \textbf{$\pm$1.13}
        & \textbf{$\pm$2.80} & \textbf{$\pm$2.58}
        & \textbf{$\pm$4.65}
        & \textbf{$\pm$2.98} & \textbf{$\pm$4.83}
        & \textbf{$\pm$6.10} \\
        \bottomrule
    \end{tabular}
\end{table}

Across the 1- and 3-shot settings, MAGIC achieves the highest MA and
FA on every dataset. It also obtains the smallest PD in all cases
except 1-shot ogbn-arxiv, where Mecoin exhibits a smaller decline but
substantially lower overall and final accuracy. The improvements are
particularly consistent on Coauthor CS and WikiCS, while the results on
the larger ogbn-arxiv graph demonstrate that MAGIC remains effective
under substantially different graph scales and structural
characteristics. Together with the 5-shot results in the main text,
these experiments show that MAGIC provides a robust balance between
novel-class adaptation and historical knowledge retention across
different supervision levels.

\subsection{Novel-Class Performance}
\label{app:novel_class_results}

We further evaluate how effectively each method learns and retains
novel classes. AN and FN denote the average novel-class accuracy across
incremental sessions and the novel-class accuracy after the final
session, respectively. Tables~\ref{tab:novel_1shot_all}--
\ref{tab:novel_5shot_all} report the results.

\begin{table}[t]
    \caption{Novel-class results (\%) under the 1-shot setting. The mean
    and standard deviation over five runs are reported in consecutive
    rows. The best result in each column is shown in bold. MAGIC Backbone:
    SGC.}
    \label{tab:novel_1shot_all}
    \centering
    \scriptsize
    \setlength{\tabcolsep}{1.5pt}
    \renewcommand{\arraystretch}{1.08}
    \begin{tabular}{L{1.70cm}*{10}{C{1.00cm}}}
        \toprule
        & \multicolumn{2}{c}{CoraFull}
        & \multicolumn{2}{c}{ogbn-arxiv}
        & \multicolumn{2}{c}{Coauthor CS}
        & \multicolumn{2}{c}{Amazon Computers}
        & \multicolumn{2}{c}{WikiCS} \\
        \cmidrule(lr){2-3}
        \cmidrule(lr){4-5}
        \cmidrule(lr){6-7}
        \cmidrule(lr){8-9}
        \cmidrule(lr){10-11}
        Method & AN & FN & AN & FN & AN & FN
        & AN & FN & AN & FN \\
        \midrule

        \rowcolor{meanrow}
        LwF
        & 16.22 & 3.27
        & 11.26 & 6.84
        & 43.70 & 22.85
        & 47.56 & 33.32
        & 46.27 & 23.89 \\
        LwF (std)
        & $\pm$2.05 & $\pm$1.27
        & $\pm$4.13 & $\pm$8.03
        & $\pm$2.48 & $\pm$9.78
        & $\pm$10.84 & $\pm$13.99
        & $\pm$4.14 & $\pm$9.25 \\

        \rowcolor{meanrow}
        EWC
        & 16.93 & 3.94
        & 13.75 & 10.75
        & 43.77 & 23.17
        & 46.60 & 33.48
        & 47.07 & 24.47 \\
        EWC (std)
        & $\pm$1.67 & $\pm$2.26
        & $\pm$3.60 & $\pm$6.83
        & $\pm$3.45 & $\pm$9.73
        & $\pm$11.08 & $\pm$13.98
        & $\pm$3.77 & $\pm$9.73 \\

        \rowcolor{meanrow}
        MAS
        & 20.32 & 8.79
        & 7.99 & 8.17
        & 58.28 & 30.91
        & 59.58 & 32.25
        & 50.77 & 26.07 \\
        MAS (std)
        & $\pm$2.48 & $\pm$3.14
        & $\pm$1.22 & $\pm$6.38
        & $\pm$8.85 & $\pm$11.04
        & $\pm$7.87 & $\pm$13.69
        & $\pm$7.63 & $\pm$8.19 \\

        \rowcolor{meanrow}
        TWP
        & 17.26 & 5.01
        & 21.31 & 11.70
        & 41.52 & 19.97
        & 46.99 & 31.85
        & 48.26 & 23.79 \\
        TWP (std)
        & $\pm$1.54 & $\pm$2.14
        & $\pm$13.05 & $\pm$6.12
        & $\pm$2.17 & $\pm$8.52
        & $\pm$5.59 & $\pm$14.49
        & $\pm$6.79 & $\pm$9.88 \\

        \rowcolor{meanrow}
        GFCIL
        & 5.48 & 5.89
        & 21.70 & 15.42
        & 12.38 & 10.35
        & 37.60 & 35.24
        & 17.03 & 17.75 \\
        GFCIL (std)
        & $\pm$0.89 & $\pm$1.13
        & $\pm$8.01 & $\pm$3.60
        & $\pm$1.78 & $\pm$3.97
        & $\pm$16.67 & $\pm$14.95
        & $\pm$6.20 & $\pm$6.45 \\

        \rowcolor{meanrow}
        Geometer
        & 8.99 & 3.94
        & 15.25 & 5.31
        & 39.97 & 27.65
        & 30.32 & 27.40
        & 26.98 & 17.21 \\
        Geometer (std)
        & $\pm$3.11 & $\pm$1.20
        & $\pm$4.82 & $\pm$5.35
        & $\pm$5.61 & $\pm$10.96
        & $\pm$9.27 & $\pm$14.82
        & $\pm$14.57 & $\pm$13.07 \\

        \rowcolor{meanrow}
        Mecoin
        & 14.04 & 11.21
        & 14.76 & 12.53
        & 32.12 & 25.31
        & 46.74 & 34.08
        & 23.03 & 19.11 \\
        Mecoin (std)
        & $\pm$3.01 & $\pm$0.79
        & $\pm$6.02 & $\pm$3.44
        & $\pm$6.48 & $\pm$4.64
        & $\pm$7.73 & $\pm$9.27
        & $\pm$9.10 & $\pm$5.70 \\

        \rowcolor{meanrow}
        TAP
        & 27.86 & 23.15
        & \textbf{35.16} & \textbf{27.92}
        & 62.02 & 59.71
        & 63.99 & 55.64
        & 55.50 & 46.74 \\
        TAP (std)
        & $\pm$2.38 & $\pm$3.36
        & \textbf{$\pm$6.53} & \textbf{$\pm$7.36}
        & $\pm$7.71 & $\pm$6.98
        & $\pm$14.66 & $\pm$10.45
        & $\pm$4.95 & $\pm$3.92 \\

        \rowcolor{meanrow}
        \textbf{MAGIC}
        & \textbf{30.19} & \textbf{28.35}
        & 31.54 & 24.23
        & \textbf{82.60} & \textbf{83.85}
        & \textbf{70.03} & \textbf{61.21}
        & \textbf{58.80} & \textbf{48.53} \\
        \textbf{MAGIC (std)}
        & \textbf{$\pm$2.70} & \textbf{$\pm$2.47}
        & $\pm$7.00 & $\pm$4.73
        & \textbf{$\pm$6.57} & \textbf{$\pm$4.66}
        & \textbf{$\pm$17.77} & \textbf{$\pm$16.90}
        & \textbf{$\pm$10.32} & \textbf{$\pm$8.95} \\
        \bottomrule
    \end{tabular}
\end{table}

\begin{table}[t]
    \caption{Novel-class results (\%) under the 3-shot setting. The mean
    and standard deviation over five runs are reported in consecutive
    rows. The best result in each column is shown in bold. MAGIC Backbone:
    SGC.}
    \label{tab:novel_3shot_all}
    \centering
    \scriptsize
    \setlength{\tabcolsep}{1.5pt}
    \renewcommand{\arraystretch}{1.08}
    \begin{tabular}{L{1.70cm}*{10}{C{1.00cm}}}
        \toprule
        & \multicolumn{2}{c}{CoraFull}
        & \multicolumn{2}{c}{ogbn-arxiv}
        & \multicolumn{2}{c}{Coauthor CS}
        & \multicolumn{2}{c}{Amazon Computers}
        & \multicolumn{2}{c}{WikiCS} \\
        \cmidrule(lr){2-3}
        \cmidrule(lr){4-5}
        \cmidrule(lr){6-7}
        \cmidrule(lr){8-9}
        \cmidrule(lr){10-11}
        Method & AN & FN & AN & FN & AN & FN
        & AN & FN & AN & FN \\
        \midrule

        \rowcolor{meanrow}
        LwF
        & 22.19 & 4.99
        & 15.97 & 6.85
        & 45.59 & 23.23
        & 57.41 & 33.81
        & 49.51 & 26.29 \\
        LwF (std)
        & $\pm$1.20 & $\pm$1.35
        & $\pm$3.74 & $\pm$8.27
        & $\pm$3.44 & $\pm$9.26
        & $\pm$12.08 & $\pm$14.12
        & $\pm$9.38 & $\pm$6.89 \\

        \rowcolor{meanrow}
        EWC
        & 22.33 & 5.10
        & 17.01 & 7.89
        & 45.86 & 23.22
        & 58.02 & 33.94
        & 50.15 & 26.40 \\
        EWC (std)
        & $\pm$1.46 & $\pm$1.80
        & $\pm$3.79 & $\pm$7.96
        & $\pm$3.43 & $\pm$9.13
        & $\pm$12.46 & $\pm$14.22
        & $\pm$9.32 & $\pm$7.20 \\

        \rowcolor{meanrow}
        MAS
        & 22.38 & 8.59
        & 9.94 & 8.05
        & 57.42 & 28.54
        & 59.77 & 31.85
        & 49.37 & 26.04 \\
        MAS (std)
        & $\pm$1.17 & $\pm$2.23
        & $\pm$1.65 & $\pm$5.87
        & $\pm$3.55 & $\pm$9.58
        & $\pm$8.10 & $\pm$14.24
        & $\pm$8.24 & $\pm$7.74 \\

        \rowcolor{meanrow}
        TWP
        & 23.33 & 7.51
        & 33.94 & 17.27
        & 46.03 & 21.53
        & 57.56 & 33.54
        & 47.12 & 26.22 \\
        TWP (std)
        & $\pm$2.21 & $\pm$1.74
        & $\pm$12.14 & $\pm$7.66
        & $\pm$2.99 & $\pm$4.42
        & $\pm$8.14 & $\pm$14.30
        & $\pm$10.40 & $\pm$7.68 \\

        \rowcolor{meanrow}
        GFCIL
        & 16.44 & 16.74
        & 35.22 & 22.86
        & 43.45 & 48.13
        & 66.11 & 57.10
        & 27.87 & 30.40 \\
        GFCIL (std)
        & $\pm$1.07 & $\pm$0.78
        & $\pm$6.80 & $\pm$2.47
        & $\pm$5.96 & $\pm$5.19
        & $\pm$12.96 & $\pm$9.68
        & $\pm$5.71 & $\pm$4.06 \\

        \rowcolor{meanrow}
        Geometer
        & 15.02 & 10.01
        & 22.15 & 12.02
        & 63.85 & 45.94
        & 43.84 & 24.95
        & 31.92 & 15.38 \\
        Geometer (std)
        & $\pm$2.07 & $\pm$1.26
        & $\pm$5.60 & $\pm$7.56
        & $\pm$3.57 & $\pm$5.96
        & $\pm$14.42 & $\pm$11.90
        & $\pm$6.13 & $\pm$4.87 \\

        \rowcolor{meanrow}
        Mecoin
        & 20.59 & 16.99
        & 23.98 & 17.86
        & 48.04 & 40.75
        & 45.71 & 36.98
        & 33.45 & 26.98 \\
        Mecoin (std)
        & $\pm$3.15 & $\pm$1.33
        & $\pm$6.32 & $\pm$2.72
        & $\pm$6.94 & $\pm$4.61
        & $\pm$5.26 & $\pm$8.29
        & $\pm$4.96 & $\pm$4.15 \\

        \rowcolor{meanrow}
        TAP
        & 38.63 & 32.19
        & 41.39 & 31.92
        & 67.97 & 64.21
        & 71.15 & 64.11
        & 61.19 & 53.76 \\
        TAP (std)
        & $\pm$3.21 & $\pm$2.15
        & $\pm$7.81 & $\pm$6.14
        & $\pm$3.05 & $\pm$3.54
        & $\pm$8.55 & $\pm$5.45
        & $\pm$9.23 & $\pm$7.55 \\

        \rowcolor{meanrow}
        \textbf{MAGIC}
        & \textbf{50.20} & \textbf{45.27}
        & \textbf{45.73} & \textbf{34.59}
        & \textbf{90.89} & \textbf{89.59}
        & \textbf{78.17} & \textbf{69.42}
        & \textbf{64.45} & \textbf{63.13} \\
        \textbf{MAGIC (std)}
        & \textbf{$\pm$1.75} & \textbf{$\pm$1.45}
        & \textbf{$\pm$6.30} & \textbf{$\pm$4.65}
        & \textbf{$\pm$1.79} & \textbf{$\pm$0.35}
        & \textbf{$\pm$9.91} & \textbf{$\pm$9.92}
        & \textbf{$\pm$9.77} & \textbf{$\pm$8.00} \\
        \bottomrule
    \end{tabular}
\end{table}

\begin{table}[t]
    \caption{Novel-class results (\%) under the 5-shot setting. The mean
    and standard deviation over five runs are reported in consecutive
    rows. The best result in each column is shown in bold. MAGIC Backbone:
    SGC.}
    \label{tab:novel_5shot_all}
    \centering
    \scriptsize
    \setlength{\tabcolsep}{1.5pt}
    \renewcommand{\arraystretch}{1.08}
    \begin{tabular}{L{1.70cm}*{10}{C{1.00cm}}}
        \toprule
        & \multicolumn{2}{c}{CoraFull}
        & \multicolumn{2}{c}{ogbn-arxiv}
        & \multicolumn{2}{c}{Coauthor CS}
        & \multicolumn{2}{c}{Amazon Computers}
        & \multicolumn{2}{c}{WikiCS} \\
        \cmidrule(lr){2-3}
        \cmidrule(lr){4-5}
        \cmidrule(lr){6-7}
        \cmidrule(lr){8-9}
        \cmidrule(lr){10-11}
        Method & AN & FN & AN & FN & AN & FN
        & AN & FN & AN & FN \\
        \midrule

        \rowcolor{meanrow}
        LwF
        & 23.86 & 5.41
        & 15.06 & 6.01
        & 46.25 & 23.51
        & 60.18 & 34.25
        & 51.46 & 27.22 \\
        LwF (std)
        & $\pm$0.88 & $\pm$1.72
        & $\pm$3.72 & $\pm$6.89
        & $\pm$4.24 & $\pm$9.79
        & $\pm$8.12 & $\pm$14.62
        & $\pm$7.31 & $\pm$8.14 \\

        \rowcolor{meanrow}
        EWC
        & 23.88 & 5.47
        & 16.26 & 7.57
        & 46.48 & 23.71
        & 59.76 & 34.16
        & 51.65 & 27.16 \\
        EWC (std)
        & $\pm$0.95 & $\pm$1.65
        & $\pm$3.72 & $\pm$8.06
        & $\pm$4.32 & $\pm$9.88
        & $\pm$7.98 & $\pm$14.46
        & $\pm$7.30 & $\pm$8.00 \\

        \rowcolor{meanrow}
        MAS
        & 23.51 & 8.60
        & 10.06 & 7.04
        & 55.78 & 28.19
        & 60.05 & 33.65
        & 49.31 & 26.31 \\
        MAS (std)
        & $\pm$1.31 & $\pm$2.18
        & $\pm$3.04 & $\pm$3.87
        & $\pm$5.72 & $\pm$9.97
        & $\pm$8.17 & $\pm$13.90
        & $\pm$7.94 & $\pm$8.63 \\

        \rowcolor{meanrow}
        TWP
        & 24.32 & 7.07
        & 32.41 & 16.16
        & 46.63 & 21.38
        & 59.14 & 33.78
        & 51.71 & 27.97 \\
        TWP (std)
        & $\pm$1.68 & $\pm$2.05
        & $\pm$9.86 & $\pm$5.62
        & $\pm$4.06 & $\pm$6.76
        & $\pm$7.71 & $\pm$14.36
        & $\pm$7.01 & $\pm$7.65 \\

        \rowcolor{meanrow}
        GFCIL
        & 25.54 & 24.70
        & 41.07 & 30.64
        & 59.95 & 63.14
        & 72.41 & 63.72
        & 45.96 & 47.45 \\
        GFCIL (std)
        & $\pm$0.47 & $\pm$1.00
        & $\pm$5.13 & $\pm$3.42
        & $\pm$4.24 & $\pm$5.04
        & $\pm$7.81 & $\pm$6.07
        & $\pm$5.43 & $\pm$2.99 \\

        \rowcolor{meanrow}
        Geometer
        & 20.86 & 15.18
        & 20.77 & 4.93
        & 71.34 & 64.10
        & 48.31 & 30.82
        & 42.64 & 22.16 \\
        Geometer (std)
        & $\pm$2.09 & $\pm$2.03
        & $\pm$4.98 & $\pm$1.51
        & $\pm$5.50 & $\pm$4.88
        & $\pm$8.58 & $\pm$11.36
        & $\pm$11.09 & $\pm$9.26 \\

        \rowcolor{meanrow}
        Mecoin
        & 25.46 & 21.66
        & 32.49 & 24.18
        & 54.25 & 44.63
        & 49.29 & 40.92
        & 43.27 & 33.42 \\
        Mecoin (std)
        & $\pm$2.12 & $\pm$1.48
        & $\pm$9.93 & $\pm$4.83
        & $\pm$4.08 & $\pm$5.25
        & $\pm$4.42 & $\pm$11.47
        & $\pm$7.36 & $\pm$4.29 \\

        \rowcolor{meanrow}
        TAP
        & 42.96 & 36.43
        & 42.77 & 33.78
        & 69.84 & 65.28
        & 72.41 & 66.94
        & 62.26 & 56.04 \\
        TAP (std)
        & $\pm$1.61 & $\pm$2.15
        & $\pm$7.90 & $\pm$6.27
        & $\pm$2.57 & $\pm$3.96
        & $\pm$8.03 & $\pm$5.88
        & $\pm$9.19 & $\pm$8.09 \\

        \rowcolor{meanrow}
        \textbf{MAGIC}
        & \textbf{56.56} & \textbf{51.81}
        & \textbf{48.71} & \textbf{37.49}
        & \textbf{92.02} & \textbf{90.87}
        & \textbf{82.27} & \textbf{73.67}
        & \textbf{69.89} & \textbf{68.81} \\
        \textbf{MAGIC (std)}
        & \textbf{$\pm$2.27} & \textbf{$\pm$2.19}
        & \textbf{$\pm$5.10} & \textbf{$\pm$5.53}
        & \textbf{$\pm$1.70} & \textbf{$\pm$0.70}
        & \textbf{$\pm$6.49} & \textbf{$\pm$10.24}
        & \textbf{$\pm$8.00} & \textbf{$\pm$5.96} \\
        \bottomrule
    \end{tabular}
\end{table}

MAGIC achieves the strongest AN and FN in every 3- and 5-shot setting,
demonstrating that it both learns novel classes effectively and
preserves them as subsequent classes arrive. Under the more challenging
1-shot setting, MAGIC remains best on four of the five datasets, while
TAP performs better on ogbn-arxiv. The advantages become particularly
clear as additional supports are provided, indicating that MAGIC
effectively combines sparse labeled evidence with transferable
topological relations. Its consistently strong FN further shows that
the gains are retained throughout the incremental process rather than
being limited to the sessions in which the classes first appear.
Together with the overall results, these findings demonstrate a
favorable balance between novel-class plasticity and continual
retention.

\subsection{Hyperparameter Sensitivity}
\label{app:hyperparameter_sensitivity}

We study the sensitivity of MAGIC on CoraFull under the 5-shot setting.
The evaluated hyperparameters include the MRF relation strength $\eta$,
pseudo-label mass $\rho$, distillation mass $\kappa$, number of prior
bins $B$, distillation temperature $T_D$, unary strength $\tau_u$, and
maximum MRF subgraph hop. We vary one parameter at a time while keeping
all other parameters at their default values. All curves report results
averaged across five runs.

\begin{figure}[t]
    \centering
    \includegraphics[width=\linewidth]{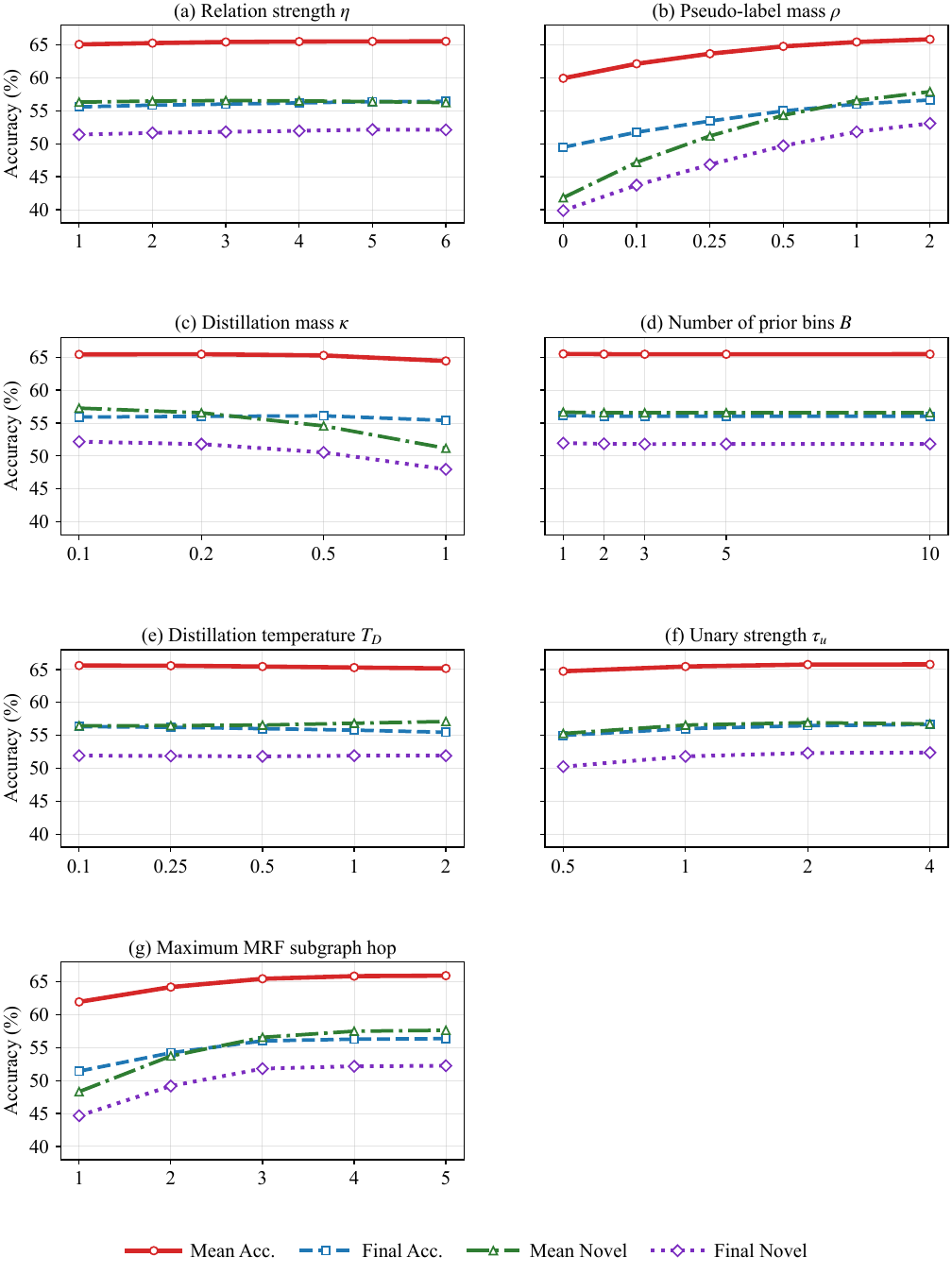}
    \caption{Hyperparameter sensitivity of MAGIC on CoraFull under the
    5-shot setting. Each subplot varies one hyperparameter while fixing
    all others to their default values. MAGIC Backbone: SGC}
    \label{fig:hyperparameter_sensitivity}
\end{figure}

Figure~\ref{fig:hyperparameter_sensitivity} shows that MAGIC is
generally robust over broad hyperparameter ranges. Increasing the
relation strength $\eta$ moderately improves overall and final
accuracy, while the novel-class metrics remain largely stable. This
indicates that stronger relational evidence is beneficial without
causing unstable topology-driven propagation. The pseudo-label mass
$\rho$ has a more pronounced effect: increasing it substantially
improves all metrics, especially novel-class accuracy, with the gains
gradually saturating at larger values. This confirms the importance of
MRF-derived auxiliary supervision while showing that its influence is
well controlled by class-balanced mass allocation.

The distillation mass $\kappa$ exhibits a clear
retention--plasticity trade-off. Moderate values preserve overall
performance, whereas excessive distillation increasingly suppresses
novel-class learning. The final overall accuracy remains relatively
stable over moderate settings but decreases when historical
supervision becomes dominant. For the distillation temperature $T_D$,
lower temperatures slightly improve overall and final accuracy by
providing sharper teacher targets, while higher temperatures produce a
small increase in mean novel-class accuracy. Final novel-class
performance is comparatively insensitive to this parameter.

MAGIC is nearly insensitive to the number of prior bins $B$, suggesting
that the class-balanced relation prior remains reliable under both
coarse and fine similarity discretizations. Increasing the unary
strength $\tau_u$ improves performance up to a moderate level, after
which the results largely saturate, demonstrating robustness to the
relative weighting of classifier evidence and pairwise relations.
Finally, expanding the MRF subgraph from one to three hops yields
substantial improvements, while the gains beyond three hops are small.
Overall, the default configuration provides a balanced compromise
between relational propagation, pseudo-label supervision, historical
retention, and novel-class adaptation.

\section{Limitations}
\label{sec:limitations}

MAGIC has two main limitations. First, it does not explicitly determine
whether the topology prior learned in the base session will have a
positive effect in each novel session. When the relation patterns of
novel classes differ substantially from those observed in the base
session, injecting the transferred prior may provide limited or even
misleading supervision. Future work could develop a session-level
prior-validity estimator that adaptively measures transferability and
adjusts the strength of prior injection.

Second, MRF inference requires access to unlabeled nodes and edges
surrounding the support set. MAGIC is therefore most applicable to
transductive or dynamically visible GFSCIL settings. Its topology-based
prior injection is more difficult to apply in strictly inductive,
privacy-constrained, or topologically incomplete scenarios, where the
relevant neighborhood structure is unavailable or only partially
observed.

\section{Reproducibility Details}
\label{app:reproducibility}

All experiments were conducted on a 64-bit Windows system
(build 10.0.26200) equipped with an NVIDIA GeForce RTX 4060 Laptop GPU
with 8\,GB of memory. The software environment used Python 3.10.20,
PyTorch 2.5.1 compiled with CUDA 12.1, PyTorch Geometric 2.8.0, and
NumPy 2.2.6. The NVIDIA driver version was 596.08. CUDA was enabled
throughout the experiments. Each result was averaged over five runs
with random seeds $\{0,1,2,3,4\}$, and we report the mean and standard
deviation across these runs.

The random seed controls class and support splits, prior-node sampling,
and the fixed random feature projection. For every dataset, the same
splits are used across the 1-, 3-, and 5-shot settings. Hyperparameters of MAGIC for each dataset in the main experiments are provided in Table~\ref{tab:dataset_hyperparameters}.

\begin{table}[t]
    \caption{Hyperparameter configurations used in the main experiments.
    The same configurations are used for the 1-, 3-, and 5-shot
    settings.}
    \label{tab:dataset_hyperparameters}
    \centering
    \scriptsize
    \setlength{\tabcolsep}{4.0pt}
    \renewcommand{\arraystretch}{1.15}
    \begin{tabular}{L{2.30cm}*{7}{C{0.90cm}}}
        \toprule
        Dataset
        & $\eta$
        & $\rho$
        & $\kappa$
        & $B$
        & $T_D$
        & $\tau_u$
        & $k$ \\
        \midrule
        CoraFull
        & 3.0 & 1.0 & 0.2 & 3  & 0.50 & 1.0 & 3 \\
        ogbn-arxiv
        & 6.0 & 1.0 & 0.5 & 10 & 0.25 & 1.0 & 3 \\
        Coauthor CS
        & 1.0 & 1.0 & 0.5 & 10 & 0.25 & 1.0 & 3 \\
        Amazon Computers
        & 1.0 & 1.0 & 0.5 & 10 & 0.25 & 1.0 & 3 \\
        WikiCS
        & 3.0 & 1.0 & 0.5 & 10 & 0.25 & 1.0 & 3 \\
        \bottomrule
    \end{tabular}
\end{table}

\end{document}

%% file: math_commands.tex
\usepackage{amsmath,amsfonts,bm}

\def\eqref#1{equation~\ref{#1}}

\def\1{\bm{1}}

\DeclareMathAlphabet{\mathsfit}{\encodingdefault}{\sfdefault}{m}{sl}
\SetMathAlphabet{\mathsfit}{bold}{\encodingdefault}{\sfdefault}{bx}{n}













%% file: iclr2027_conference.bbl
\begin{thebibliography}{38}
\providecommand{\natexlab}[1]{#1}
\providecommand{\url}[1]{\texttt{#1}}
\expandafter\ifx\csname urlstyle\endcsname\relax
  \providecommand{\doi}[1]{doi: #1}\else
  \providecommand{\doi}{doi: \begingroup \urlstyle{rm}\Url}\fi

\bibitem[Aljundi et~al.(2018)Aljundi, Babiloni, Elhoseiny, Rohrbach, and Tuytelaars]{aljundi2018memory}
Rahaf Aljundi, Francesca Babiloni, Mohamed Elhoseiny, Marcus Rohrbach, and Tinne Tuytelaars.
\newblock Memory aware synapses: Learning what (not) to forget.
\newblock In \emph{European Conference on Computer Vision}, pp.\  139--154, 2018.

\bibitem[Bojchevski \& G{\"u}nnemann(2018)Bojchevski and G{\"u}nnemann]{bojchevski2018deep}
Aleksandar Bojchevski and Stephan G{\"u}nnemann.
\newblock Deep gaussian embedding of graphs: Unsupervised inductive learning via ranking.
\newblock In \emph{International Conference on Learning Representations}, 2018.

\bibitem[Chi et~al.(2022)Chi, Gu, Liu, Wang, Yu, and Tang]{chi2022metafscil}
Zhixiang Chi, Li~Gu, Huan Liu, Yang Wang, Yuanhao Yu, and Jin Tang.
\newblock {MetaFSCIL}: A meta-learning approach for few-shot class incremental learning.
\newblock In \emph{Proceedings of the IEEE/CVF Conference on Computer Vision and Pattern Recognition}, pp.\  14166--14175, 2022.

\bibitem[Dempster et~al.(1977)Dempster, Laird, and Rubin]{dempster1977maximum}
Arthur~P. Dempster, Nan~M. Laird, and Donald~B. Rubin.
\newblock Maximum likelihood from incomplete data via the {EM} algorithm.
\newblock \emph{Journal of the Royal Statistical Society: Series B}, 39\penalty0 (1):\penalty0 1--22, 1977.

\bibitem[Gasteiger et~al.(2019)Gasteiger, Bojchevski, and G{\"u}nnemann]{gasteiger2019ppnp}
Johannes Gasteiger, Aleksandar Bojchevski, and Stephan G{\"u}nnemann.
\newblock Predict then propagate: Graph neural networks meet personalized pagerank.
\newblock In \emph{International Conference on Learning Representations}, 2019.

\bibitem[Guo et~al.(2025)Guo, Sun, Zhang, Yuan, Zhuang, Fu, and Li]{guo2025graphkeeper}
Zihao Guo, Qingyun Sun, Ziwei Zhang, Haonan Yuan, Huiping Zhuang, Xingcheng Fu, and Jianxin Li.
\newblock {GraphKeeper}: Graph domain-incremental learning via knowledge disentanglement and preservation.
\newblock In \emph{Advances in Neural Information Processing Systems}, volume~38, 2025.

\bibitem[Hamilton et~al.(2017)Hamilton, Ying, and Leskovec]{hamilton2017inductive}
William~L. Hamilton, Rex Ying, and Jure Leskovec.
\newblock Inductive representation learning on large graphs.
\newblock In \emph{Advances in Neural Information Processing Systems}, volume~30, 2017.

\bibitem[Hinton et~al.(2015)Hinton, Vinyals, and Dean]{hinton2015distilling}
Geoffrey Hinton, Oriol Vinyals, and Jeff Dean.
\newblock Distilling the knowledge in a neural network.
\newblock \emph{arXiv preprint arXiv:1503.02531}, 2015.

\bibitem[Hu et~al.(2020)Hu, Fey, Zitnik, Dong, Ren, Liu, Catasta, and Leskovec]{hu2020ogb}
Weihua Hu, Matthias Fey, Marinka Zitnik, Yuxiao Dong, Hongyu Ren, Bowen Liu, Michele Catasta, and Jure Leskovec.
\newblock Open graph benchmark: Datasets for machine learning on graphs.
\newblock In \emph{Advances in Neural Information Processing Systems}, volume~33, pp.\  22118--22133, 2020.

\bibitem[Kim et~al.(2022)Kim, Yun, and Kang]{kim2022dygrain}
Seoyoon Kim, Seongjun Yun, and Jaewoo Kang.
\newblock {DyGRAIN}: An incremental learning framework for dynamic graphs.
\newblock In \emph{Proceedings of the Thirty-First International Joint Conference on Artificial Intelligence}, pp.\  3157--3163, 2022.
\newblock \doi{10.24963/ijcai.2022/438}.

\bibitem[Kipf \& Welling(2017)Kipf and Welling]{kipf2017gcn}
Thomas~N. Kipf and Max Welling.
\newblock Semi-supervised classification with graph convolutional networks.
\newblock In \emph{International Conference on Learning Representations}, 2017.

\bibitem[Kirkpatrick et~al.(2017)Kirkpatrick, Pascanu, Rabinowitz, Veness, Desjardins, Rusu, Milan, Quan, Ramalho, Grabska-Barwinska, Hassabis, Clopath, Kumaran, and Hadsell]{kirkpatrick2017overcoming}
James Kirkpatrick, Razvan Pascanu, Neil Rabinowitz, Joel Veness, Guillaume Desjardins, Andrei~A. Rusu, Kieran Milan, John Quan, Tiago Ramalho, Agnieszka Grabska-Barwinska, Demis Hassabis, Claudia Clopath, Dharshan Kumaran, and Raia Hadsell.
\newblock Overcoming catastrophic forgetting in neural networks.
\newblock \emph{Proceedings of the National Academy of Sciences}, 114\penalty0 (13):\penalty0 3521--3526, 2017.
\newblock \doi{10.1073/pnas.1611835114}.

\bibitem[Lee et~al.(2026)Lee, Kong, Jeong, and Yoo]{lee2026rgvt}
Dooho Lee, Myeong Kong, Minho Jeong, and Jaemin Yoo.
\newblock View space: Learning representation across arbitrary graphs.
\newblock In \emph{International Conference on Machine Learning}, 2026.

\bibitem[Li et~al.(2024)Li, Zhang, Gao, and Qi]{li2024mecoin}
Dong Li, Aijia Zhang, Junqi Gao, and Biqing Qi.
\newblock An efficient memory module for graph few-shot class-incremental learning.
\newblock In \emph{Advances in Neural Information Processing Systems}, volume~37, 2024.

\bibitem[Li et~al.(2025)Li, Moghadam, Peng, Ye, and Koniusz]{li2025inductive}
Yayong Li, Peyman Moghadam, Can Peng, Nan Ye, and Piotr Koniusz.
\newblock Inductive graph few-shot class incremental learning.
\newblock In \emph{Proceedings of the Eighteenth ACM International Conference on Web Search and Data Mining}, pp.\  466--474, 2025.
\newblock \doi{10.1145/3701551.3703578}.

\bibitem[Li \& Hoiem(2016)Li and Hoiem]{li2016learning}
Zhizhong Li and Derek Hoiem.
\newblock Learning without forgetting.
\newblock In \emph{European Conference on Computer Vision}, pp.\  614--629. Springer, 2016.
\newblock \doi{10.1007/978-3-319-46493-0_37}.

\bibitem[Liu et~al.(2021)Liu, Yang, and Wang]{liu2021twp}
Huihui Liu, Yiding Yang, and Xinchao Wang.
\newblock Overcoming catastrophic forgetting in graph neural networks.
\newblock \emph{Proceedings of the AAAI Conference on Artificial Intelligence}, 35\penalty0 (10):\penalty0 8653--8661, 2021.
\newblock \doi{10.1609/aaai.v35i10.17049}.

\bibitem[Liu et~al.(2023)Liu, Qiu, and Huang]{liu2023cat}
Yilun Liu, Ruihong Qiu, and Zi~Huang.
\newblock {CaT}: Balanced continual graph learning with graph condensation.
\newblock In \emph{IEEE International Conference on Data Mining}, pp.\  1157--1162, 2023.
\newblock \doi{10.1109/ICDM58522.2023.00141}.

\bibitem[Liu et~al.(2025)Liu, Qiu, Tang, Yin, and Huang]{liu2025puma}
Yilun Liu, Ruihong Qiu, Yanran Tang, Hongzhi Yin, and Zi~Huang.
\newblock {PUMA}: Efficient continual graph learning for node classification with graph condensation.
\newblock \emph{IEEE Transactions on Knowledge and Data Engineering}, 37\penalty0 (1):\penalty0 449--461, 2025.
\newblock \doi{10.1109/TKDE.2024.3485691}.

\bibitem[Lu et~al.(2022)Lu, Gan, Yang, Zhang, Fu, and Wang]{lu2022geometer}
Bin Lu, Xiaoying Gan, Lina Yang, Weinan Zhang, Luoyi Fu, and Xinbing Wang.
\newblock Geometer: Graph few-shot class-incremental learning via prototype representation.
\newblock In \emph{Proceedings of the 28th ACM SIGKDD Conference on Knowledge Discovery and Data Mining}, pp.\  1152--1161, 2022.
\newblock \doi{10.1145/3534678.3539280}.

\bibitem[Mernyei \& Cangea(2020)Mernyei and Cangea]{mernyei2020wikics}
P{\'e}ter Mernyei and C{\u{a}}t{\u{a}}lina Cangea.
\newblock {Wiki-CS}: A wikipedia-based benchmark for graph neural networks.
\newblock \emph{arXiv preprint arXiv:2007.02901}, 2020.

\bibitem[Niu et~al.(2024)Niu, Pang, and Chen]{niu2024delome}
Chaoxi Niu, Guansong Pang, and Ling Chen.
\newblock Graph continual learning with debiased lossless memory replay.
\newblock In \emph{Proceedings of the 27th European Conference on Artificial Intelligence}, volume 392, pp.\  1808--1815. IOS Press, 2024.
\newblock \doi{10.3233/FAIA240692}.

\bibitem[Shchur et~al.(2018)Shchur, Mumme, Bojchevski, and G{\"u}nnemann]{shchur2018pitfalls}
Oleksandr Shchur, Maximilian Mumme, Aleksandar Bojchevski, and Stephan G{\"u}nnemann.
\newblock Pitfalls of graph neural network evaluation.
\newblock \emph{arXiv preprint arXiv:1811.05868}, 2018.

\bibitem[Tan et~al.(2022)Tan, Ding, Guo, and Liu]{tan2022gfcil}
Zhen Tan, Kaize Ding, Ruocheng Guo, and Huan Liu.
\newblock Graph few-shot class-incremental learning.
\newblock In \emph{Proceedings of the Fifteenth ACM International Conference on Web Search and Data Mining}, pp.\  987--996, 2022.
\newblock \doi{10.1145/3488560.3498455}.

\bibitem[Tao et~al.(2020)Tao, Hong, Chang, Dong, Wei, and Gong]{tao2020fewshot}
Xiaoyu Tao, Xiaopeng Hong, Xinyuan Chang, Songlin Dong, Xing Wei, and Yihong Gong.
\newblock Few-shot class-incremental learning.
\newblock In \emph{Proceedings of the IEEE/CVF Conference on Computer Vision and Pattern Recognition}, pp.\  12183--12192, 2020.

\bibitem[Veli\v{c}kovi\'{c} et~al.(2018)Veli\v{c}kovi\'{c}, Cucurull, Casanova, Romero, Li\`{o}, and Bengio]{velickovic2018graph}
Petar Veli\v{c}kovi\'{c}, Guillem Cucurull, Arantxa Casanova, Adriana Romero, Pietro Li\`{o}, and Yoshua Bengio.
\newblock Graph attention networks.
\newblock In \emph{International Conference on Learning Representations}, 2018.

\bibitem[Wainwright \& Jordan(2008)Wainwright and Jordan]{wainwright2008graphical}
Martin~J. Wainwright and Michael~I. Jordan.
\newblock Graphical models, exponential families, and variational inference.
\newblock \emph{Foundations and Trends in Machine Learning}, 1\penalty0 (1--2):\penalty0 1--305, 2008.
\newblock \doi{10.1561/2200000001}.

\bibitem[Wang et~al.(2021)Wang, Jia, and Gong]{wang2021pmrf}
Binghui Wang, Jinyuan Jia, and Neil~Zhenqiang Gong.
\newblock Semi-supervised node classification on graphs: Markov random fields vs. graph neural networks.
\newblock \emph{Proceedings of the AAAI Conference on Artificial Intelligence}, 35\penalty0 (11):\penalty0 10093--10101, 2021.
\newblock \doi{10.1609/aaai.v35i11.17211}.

\bibitem[Wu et~al.(2019)Wu, Souza, Zhang, Fifty, Yu, and Weinberger]{wu2019sgc}
Felix Wu, Amauri Souza, Tianyi Zhang, Christopher Fifty, Tao Yu, and Kilian Weinberger.
\newblock Simplifying graph convolutional networks.
\newblock In \emph{Proceedings of the 36th International Conference on Machine Learning}, volume~97, pp.\  6861--6871. PMLR, 2019.

\bibitem[Xia \& Huang(2026)Xia and Huang]{xia2026anygraph}
Lianghao Xia and Chao Huang.
\newblock {AnyGraph}: Graph foundation model in the wild.
\newblock In \emph{Findings of the Association for Computational Linguistics: ACL 2026}, pp.\  882--896. Association for Computational Linguistics, 2026.
\newblock \doi{10.18653/v1/2026.findings-acl.44}.

\bibitem[Xu et~al.(2019)Xu, Shen, Cao, Cen, and Cheng]{xu2019graphheat}
Bingbing Xu, Huawei Shen, Qi~Cao, Keting Cen, and Xueqi Cheng.
\newblock Graph convolutional networks using heat kernel for semi-supervised learning.
\newblock In \emph{Proceedings of the Twenty-Eighth International Joint Conference on Artificial Intelligence}, pp.\  1928--1934. International Joint Conferences on Artificial Intelligence Organization, 2019.
\newblock \doi{10.24963/ijcai.2019/267}.

\bibitem[Zhang et~al.(2021)Zhang, Song, Lin, Zheng, Pan, and Xu]{zhang2021cec}
Chi Zhang, Nan Song, Guosheng Lin, Yun Zheng, Pan Pan, and Yinghui Xu.
\newblock Few-shot incremental learning with continually evolved classifiers.
\newblock In \emph{Proceedings of the IEEE/CVF Conference on Computer Vision and Pattern Recognition}, pp.\  12455--12464, 2021.

\bibitem[Zhang et~al.(2022)Zhang, Song, and Tao]{zhang2022cglb}
Xikun Zhang, Dongjin Song, and Dacheng Tao.
\newblock {CGLB}: Benchmark tasks for continual graph learning.
\newblock In \emph{Advances in Neural Information Processing Systems, Datasets and Benchmarks Track}, volume~35, 2022.

\bibitem[Zhou et~al.(2022)Zhou, Wang, Ye, Ma, Pu, and Zhan]{zhou2022fact}
Da-Wei Zhou, Fu-Yun Wang, Han-Jia Ye, Liang Ma, Shiliang Pu, and De-Chuan Zhan.
\newblock Forward compatible few-shot class-incremental learning.
\newblock In \emph{Proceedings of the IEEE/CVF Conference on Computer Vision and Pattern Recognition}, pp.\  9046--9056, 2022.

\bibitem[Zhou \& Cao(2021)Zhou and Cao]{zhou2021ergnn}
Fan Zhou and Chengtai Cao.
\newblock Overcoming catastrophic forgetting in graph neural networks with experience replay.
\newblock \emph{Proceedings of the AAAI Conference on Artificial Intelligence}, 35\penalty0 (5):\penalty0 4714--4722, 2021.
\newblock \doi{10.1609/aaai.v35i5.16602}.

\bibitem[Zhu \& Koniusz(2021)Zhu and Koniusz]{zhu2021ssgc}
Hao Zhu and Piotr Koniusz.
\newblock Simple spectral graph convolution.
\newblock In \emph{International Conference on Learning Representations}, 2021.

\bibitem[Zhuang et~al.(2022)Zhuang, Weng, Wei, Xie, Toh, and Lin]{zhuang2022acil}
Huiping Zhuang, Zhenyu Weng, Hongxin Wei, Renchunzi Xie, Kar-Ann Toh, and Zhiping Lin.
\newblock {ACIL}: Analytic class-incremental learning with absolute memorization and privacy protection.
\newblock In \emph{Advances in Neural Information Processing Systems}, volume~35, pp.\  11602--11614, 2022.

\bibitem[Zhuang et~al.(2023)Zhuang, Weng, He, Lin, and Zeng]{zhuang2023gkeal}
Huiping Zhuang, Zhenyu Weng, Run He, Zhiping Lin, and Ziqian Zeng.
\newblock {GKEAL}: Gaussian kernel embedded analytic learning for few-shot class incremental task.
\newblock In \emph{Proceedings of the IEEE/CVF Conference on Computer Vision and Pattern Recognition}, pp.\  7746--7755, 2023.
\newblock \doi{10.1109/CVPR52729.2023.00748}.

\end{thebibliography}
